\documentclass[twoside]{article}
\usepackage[accepted]{aistats2019}
\usepackage{amsmath,amssymb,amsthm}
\usepackage{natbib}
\usepackage{algorithm, algpseudocode}

\usepackage{multirow}

\usepackage{hyperref}
\hypersetup{
    colorlinks = true,
	citecolor=blue,
	linkcolor=blue
}
\usepackage{subfigure}
\usepackage{framed}
\usepackage{graphicx}

\allowdisplaybreaks[3]

\usepackage{amscd}

\newcommand{\bs}{\boldsymbol}

\DeclareMathOperator*{\argmin}{arg\,min}

\theoremstyle{definition} %% 定理環境を斜字から通常の字体に戻す
\newtheorem{theo}{Theorem}[section]
\newtheorem{lem}[theo]{Lemma}

\usepackage{hyperref}       % hyperlinks
\usepackage{url}            % simple URL typesetting
\usepackage{amsfonts}       % blackboard math symbols
\usepackage{nicefrac}       % compact symbols for 1/2, etc.
\usepackage{enumerate}
\usepackage{authblk}

\begin{document}



\twocolumn[
\aistatstitle{Robust Graph Embedding with Noisy Link Weights}

\aistatsauthor{Akifumi Okuno$^{\dagger,\ddagger}$ \\ \url{okuno@sys.i.kyoto-u.ac.jp} \And Hidetoshi Shimodaira$^{\dagger,\ddagger}$ \\ \url{shimo@i.kyoto-u.ac.jp}}
\aistatsaddress{
$^{\dagger}$Graduate School of Informatics, Kyoto University, \quad  $^{\ddagger}$RIKEN Center for Artificial Intelligence Project~(AIP)} 
]

% \twocolumn[
% \aistatstitle{Robust Graph Embedding with Noisy Link Weights}

% \aistatsauthor{Anonymous author(s)}
% \aistatsaddress{Anonymous institution(s)} 
% ]

% アブストでは省略形なしにしておきます
\begin{abstract}
We propose \textit{$\beta$-graph embedding} for robustly learning feature vectors from data vectors and noisy link weights.
A newly introduced \textit{empirical moment $\beta$-score} reduces the influence of contamination and robustly measures the difference between the underlying correct expected weights of links and the specified generative model.
The proposed method is computationally tractable; we employ a minibatch-based efficient stochastic algorithm and prove that this algorithm locally minimizes the empirical moment $\beta$-score.
We conduct numerical experiments on synthetic and real-world datasets. 
\end{abstract}

\section{INTRODUCTION}

%% グラフ埋め込みの重要性
In the past few decades, graph embedding~(GE) that learns \textit{feature vectors} of given graph nodes had high level of demand in a broad range of fields. 
Just an example among many, embedding social networks whose nodes and link weights represent users and their relationships, respectively, produces user feature vectors. 
% and is required to produce user feature vectors. 
Traditional multivariate statistical methods, such as clustering and classification, can then be applied to these feature vectors~\citep{yan2007graph,goyal2018graph}, whereas these analysis methods, in general, cannot be applied directly to unprocessed graph nodes. 

%Given a graph, understanding the interaction between node pairs and link weights is highly demanded in a broad range of fields these days. 
%To mention but a few, analysis of social networking services~(SNS) has gathered a great interest~\citep{hanneman2005introduction}. 
%Understanding their interactions representing whether user pairs~(represented by node pairs) know each other~(represented by link weights) enables us to infer which pair of users is supposed to be associated even when their relationship is preliminary unknown. 

%% classical GEとLPPの説明，ただし計算量が大きい
Classical GE typified by spectral GE~\citep[SGE; see][]{chung1997spectral,belkin2001laplacian} computes feature vectors so that their inner product similarities represent the link weights, and the locality-preserving projections~\citep[LPP; see][]{he2004locality} extends SGE so that it additionally considers pre-obtained data vectors of nodes. 
LPP computes feature vectors by linear transformation of data vectors; the local configuration of vectors is partially preserved through the transformation. 
Although SGE and LPP experimentally demonstrate reasonable performance, their computational complexity is high owing to eigendecomposition.

%% 尤度ベースのGEが提案されて，NN-basedのnon-linear extensionが最近研究サれている
To reduce the high computational complexity, 
\citet{tang2015line} proposed a computationally efficient GE named large-scale information network embedding~(LINE), 
which is based on the stochastic maximization of the likelihood over link weights. 
Although LINE, as well as SGE, does not consider pre-obtained data vectors of nodes, the graph embedding can be extended to utilize the data vectors by incorporating neural networks~\citep{wang2016structural,kipf2017semi,dai2018adversarial}. 
The graph embedding can be further extended to deal with multi-view setting
\citep{okuno2018probabilistic}, where each vector is assigned to one of multiple types of vectors (e.g. image vectors, text vectors) with different dimensionalities.
%The inner product similarity between two neural network-based feature vectors is especially known as Siamese-style similarity~\citep{bromley1994signature}. 

%\textbf{ノードの特徴量を活用することでgeometricな情報を盛り込める．} 
%Classicalなspectral graph埋め込みなどがgraph embeddingを利用していた一方で，
%nodeの特徴量を活用することでgeometricな情報をfeature vectorに盛り込めることが分かってきた．
%here, 各nodeがpreliminarly-obtained data vector representationを持っている状況を考える．
%例えばLocality Preserving Projections~\citep{he2004locality}はfeature vectorsをdata vectorの線形変換によって定義することで，feature vectorsにdata vectorsのgeometricな情報をダイレクトに反映させた．
%一方で，線形のモデルは表現力が弱すぎるので非線形モデルとしてneural networkを利用するモデルが近年提案された
%~\citep{wang2016structural,kipf2017semi,hamilton2017inductive}. 

\vspace{-0.5em}

\begin{figure}[htbp]
\centering
\subfigure[$\beta=0$]{
		\includegraphics[width=2.3cm]{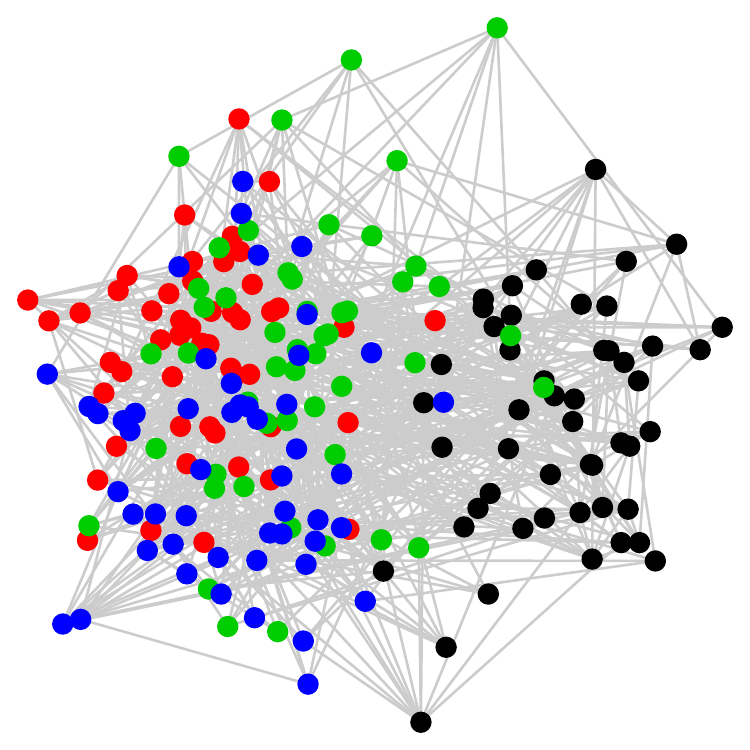}
	\label{fig:likelihood}
}
\subfigure[$\beta=0.5$]{
		\includegraphics[width=2.3cm]{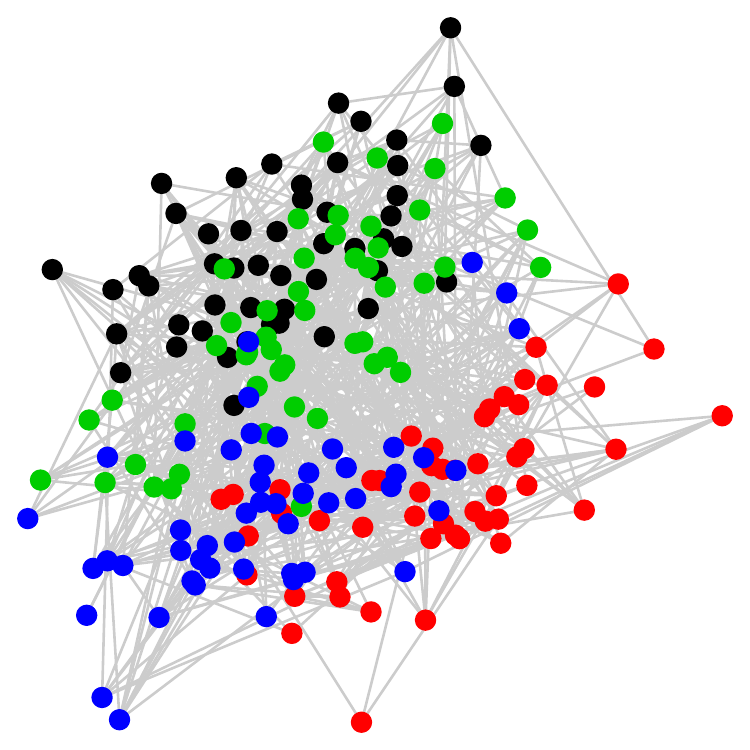}
	\label{fig:beta05}
}
\subfigure[$\beta=1.0$]{
\includegraphics[width=2.3cm]{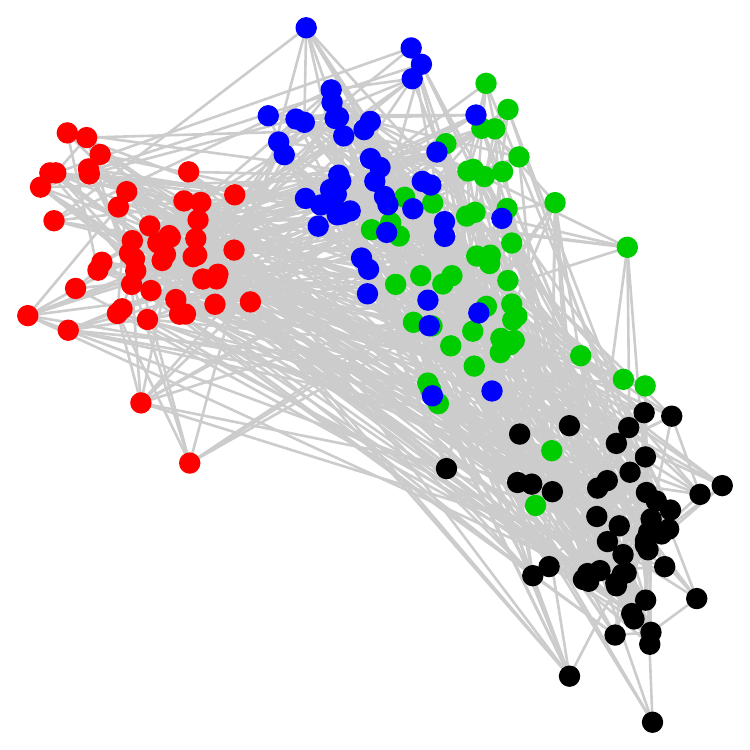}
	\label{fig:beta1}}
\vspace{-0.5em}
\caption{
\small 
\textbf{Proposed $\beta$-GE applied to noisy graph with four clusters.
\subref{fig:likelihood}~corresponds to the existing likelihood-based GE.
\subref{fig:beta1}~reduces the influence of contamination in noisy link weights.
} 
See Section~\ref{subsec:numeraical_experiment} for detailed settings. 
}
\label{fig:illustrative_example}
\end{figure}

%% ToDo: ここをふやす　
Although these GEs have been successful in many applications, their performance relies on the quality of observed link weights. 
However, these observed link weights, in practice, may contain noise. 
%For instance, 
We especially call the link weights with noise as \textit{noisy link weights}. 
As shown in Figure~\ref{fig:likelihood}, the noise may degrade the GE's performance; 
%since existing GE methods rely on the precision of observed links, 
existing GE is hard to recover the underlying cluster structure from noisy link weights. 
To overcome this problem, GE should be robustified.

In order to capture the underlying link structure through the noisy link weights, noise-tolerant loss functions can be used. 
Although a variety of noise-tolerant loss functions have been considered in robust statistics,
there have been only a few attempts at using such a noise-tolerant loss function in GE.
To the best of the authors' knowledge, only \citet{pang2010outlier} and \citet{zhang2014robust} have incorporated noise-tolerant $\ell_1$- and $\ell_{2,1}$-norm-based loss functions into classical GE. 
Although these particular types of loss functions are experimentally demonstrated to alleviate the adverse effect of outlier data vectors, they are not theoretically guaranteed to be robust against noisy link weights.

To obtain theoretically-guaranteed robust GE, we attempt to employ $\beta$-score, that robustly measures the difference between two non-negative functions. 
$\beta$-score is also called as density-power score~\citep{kanamori2014affine,kanamori2015robust}, and it is especially called as $\beta$-cross entropy~\citep[see, e.g.][]{futami2018variational} if the two non-negative functions are limited to probability density function (pdf) or probability mass function (pmf). 
It is well known that Kullback-Leibler divergence is expressed as the difference of two cross-entropies ($\beta=0$),
and quite similarly $\beta$-divergence~\citep{cichocki2010families}, which is also known as density-power divergence~\citep{basu1998robust}, is expressed as the difference of two $\beta$-cross entropies.

By specifying Poisson distribution for link weights, a probabilistic generative model of GE has been considered~\citep{okuno2018probabilistic}, and $\beta$-score can be applied to this model.
However, naively applying $\beta$-score to the probabilistic generative model of GE has two disadvantages: $\beta$-score in this setting is 
(i) computationally intractable, and 
(ii) sensitive to distributional misspecification of the generative model, because 
$\beta$-score measures the difference between two probabilistic models.

%\textbf{そこで，noise-torrelatntな損失を用いた学習が代わりに用いられる}
%特に，通常の2値判別の問題でnoise labelにtorrelantな損失を利用する方法が\citep{mnih2012learning,xiao2015learning,patrini2017making,tanaka2018joint}で提案されているが，
%理論保証がない．
%\citet{natarajan2013learning}がnoise contamination ratioがknownの限定的な状況で，
%\citet{ghosh2017robust}がnoiseがuniformである状況で
%それぞれnoiseにtorelantであることがtheoretically guaranteedな方法を提案したけど仮定が強い．
%また，これらは各ベクトル$\bs x_i$にラベル$y_i$がついているsingle-indexの問題設定を扱っていて，
%ペア$(\bs x_i,\bs x_j)$ごとにラベル$w_{ij}$がついているグラフ埋め込みを扱うことができなかった．
%グラフ埋め込みの文脈では，損失関数にsquare-lossではなく
%$\ell_1$-や$\ell_{2,1}$-lossを使ったりするやつがある~\citep{pang2010outlier,zhang2014robust}が理論保証が与えられてない

In order to avoid these two disadvantages, we introduce \emph{moment $\beta$-score} that measures the difference in terms of the expected values instead of probability distributions of link weights.
The moment $\beta$-score is estimated by  \textit{empirical moment $\beta$-score (EMBS)}, which is robust against noise in link weights, 
and is also carefully designed to be 
(i) computationally tractable, 
and 
(ii) free from the distributional specification.
% The property (ii) follows from the fact that EMBS converges in probability to UBSF regarding the \emph{expectation of link weights}, instead of the probability model of link weights.
By incorporating EMBS into existing likelihood-based GE using the probabilistic generative model of link weights, we propose a theoretically proven robust GE named \textit{$\beta$-graph embedding ($\beta$-GE)}. 
This new method naturally extends the likelihood-based GE because the EMBS reduces to the negative log-likelihood of link weights as $\beta \downarrow 0$. 

Our contribution is summarized as follows.

\begin{enumerate}[{(1)}]
\item We propose a novel EMBS, that is a robust score function against noise in link weights as shown in Theorem~\ref{theo:div_perturbation}. 
EMBS is also computationally tractable and it is free from the distributional specification. 
\item We propose $\beta$-GE by simply incorporating the newly proposed EMBS into existing likelihood-based GE (Section~\ref{subsec:beta_estimation}). 
%As EMBS is robust against noises in link weights, $\beta$-GE is a robustification of the likelihood-based GE. 
As far as we know, $\beta$-GE is the first GE whose robustness is theoretically shown. 
\item We propose an efficient minibatch-based stochastic algorithm, which can be proved to locally minimize the EMBS  (Section~\ref{subsec:minibatch_SGD}).

\item We conduct experiments on both synthetic and real-world datasets (Section~\ref{sec:experiments}).
\end{enumerate}

%The remainder of this paper is organized as follows. 
%In Section~\ref{subsec:graph_embedding}, we review existing likelihood-based GE. 
%In Section~\ref{sec:robust_graph_embedding}, we propose EMBS and prove its robustness against noise in link weights. Then we propose \textit{$\beta$-GE} by incorporating EMBS into the likelihood-based GE. 
%An efficient minibatch-based stochastic algorithm is provided to minimize the EMBS, and we prove that this algorithm can exactly locally minimize the EMBS in probability. 
% In Section~\ref{sec:related_works}, we review related works. 
%In , we conduct experiments on both synthetic and real-world datasets. 
%In Section~\ref{sec:conclusion}, we conclude this paper. 

\section{LIKELIHOOD-BASED GRAPH EMBEDDING}
%\subsection{Likelihood-based graph embedding}
\label{subsec:graph_embedding}
In this section, we review existing likelihood-based GE. 
Throughout this paper, we consider a setting similar to that presented in~\citet{okuno2018probabilistic}. 
Although only single-view graph embedding is considered below, its extension to multi-view setting is straightforward as explained in Section~\ref{subsec:multi_view}.

Our dataset consists of data vectors $\{\bs x_i\}_{i=1}^{n}$ and link weights $\{w_{ij}\}_{(i,j) \in \mathcal{I}_n}$, where $\mathcal{I}_n:=\{(i,j) \in \mathbb{N}^2 \mid 1 \leq i <j \leq n\}$.
The $p$-dimensional data vector $\bs x_i\in \mathcal{X}$ takes values in a compact set $\mathcal{X} \subset \mathbb{R}^p$.
The link weight $w_{ij} \in \{0,1,2,\ldots\}$ is non-negative integer, and it represents the strength of association between $\bs x_i$ and $\bs x_j$.

Similarly to existing methods~\citep{perozzi2014deepwalk,tang2015line,grover2016node2vec,kipf2016variational}, which employ Bernoulli-based models for considering binary weights, \citet{okuno2018probabilistic} employs a Poisson-based model
\begin{align}
	w_{ij} \mid \bs x_i,\bs x_j
	&\overset{\text{indep.}}{\sim}
	\text{Po}(\mu_{\bs \theta}(\bs x_i,\bs x_j))
	\label{eq:w_model}  
\end{align}
by taking $w_{ij}$ as random variables for all $(i,j) \in \mathcal{I}_n$. 
Note that the model (\ref{eq:w_model}) does not restrict weights $\{w_{ij}\}$ to be binary.
The conditional expected value $E(w_{ij} | \bs x_i,\bs x_j   )$ is specified by a symmetric, continuous, and non-negative function $\mu_{\bs \theta}(\bs x_i,\bs x_j)$, which represents a similarity between two data vectors $\bs x_i,\bs x_j \in \mathcal{X}$, where  $\bs \theta \in \bs \Theta$ is a parameter vector and $\bs \Theta \subset \mathbb{R}^q$ is a non-empty compact set.
% As $\mu_{\bs \theta}(\bs x_i,\bs x_j)$ is the conditional expectation of $w_{ij}$ given $(\bs x_i,\bs x_j)$ in (\ref{eq:w_model}), we call the model an \textit{expected weight model}. 
Although any symmetric, continuous, and non-negative function can be used as the expected weight function, in this paper, the function is specified as
\begin{align}
\log \mu_{\bs \theta}(\bs x_i,\bs x_j)
:=
\langle f_{\bs \psi}(\bs x_i),f_{\bs \psi}(\bs x_j) \rangle-\gamma, 
\label{eq:csips_model}
\end{align}
where $f_{\bs \psi}:\mathbb{R}^p \to \mathbb{R}^K$ is arbitrary continuous map whose parameter vector is $\bs \psi$, $\gamma \in \mathbb{R}$ is a scaling parameter regularizing the sparseness of $\{w_{ij}\}$, and $\bs \theta$ denotes $(\bs \psi,\gamma)$. 
Typically, a NN or linear transformation is used as $f_{\bs \psi}$, and the \textit{inner product similarity} (IPS) between two NNs $\langle f_{\bs \psi}(\bs x_i),f_{\bs \psi}(\bs x_j) \rangle$ is also known as siamese-style similarity~\citep{bromley1994signature}. 

IPS is used for a wide variety of GEs~\citep{wang2016structural,kipf2016variational,
hamilton2017inductive}, and has been proven to approximate arbitrary positive-definite~(PD) similarities~\citep[Theorem 5.1]{okuno2018probabilistic}. 
In addition, equation (\ref{eq:csips_model}), which incorporates a new parameter $\gamma \in \mathbb{R}$ into IPS, is specifically called \textit{constantly-shifted IPS} (C-SIPS), and is proved to approximate conditionally PD similarities, a wider class of similarities than PD similarities~\citep[Theorem 4.2]{okuno2019graph}. 
Thus, C-SIPS has a sufficiently high representation capability.

%Especially, (\ref{eq:csips_model}) is called as Constantly-Shifted Inner Product Similarity~(C-SIPS), and is proved to approximate any conditionally positive definite~(CPD) similarities arbitrary well when considering NN with sufficiently large number of output units and hidden units~\citep{okuno2018representation}. 
%Since CPD is a vast class of similarities, which includes Poincare and Wasserstein distance for example, C-SIPS has a remarkably high representation capability. 

Once the optimal $\bs{\hat \theta}=(\bs{\hat \psi},\hat{\gamma})$ is obtained, data vectors $\{\bs x_i\}$ can be transformed to \textit{feature vectors} $\{\bs y_i\}$ as
% transform data vector $\bs x_i$ to the corresponding feature vector $\bs y_i$, as
\begin{align}
	\mathbb{R}^p
	\supset
	\mathcal{X} \ni \bs x_i
	\mapsto
	\bs y_i:=f_{\bs{\hat \psi}}(\bs x_i) \in \mathbb{R}^K
	\label{eq:transformation}
\end{align}
for all $i=1,2,\ldots,n$. 
The IPS $\langle \bs y_i,\bs y_j \rangle$ of obtained feature vectors $\{\bs y_i\}$ indicates the similarity between data vectors $\bs x_i,\bs x_j$. 
This transformation usually reduces the dimensionality of data vectors from $p$ to $K (\leq p)$. 
Thus, the obtained vectors are expected to reduce their redundancy while considering the information on link weights. 
%Furthermore, even the similarity between newly-obtained data vectors $\bs x_{n+1},\bs x_{n+2}$, whose link weight $w_{n+1,n+2}$ is unobserved, can be predicted through the inner product $\langle \bs y_{n+1},\bs y_{n+2} \rangle$. 
%; applying existing statistical methods such as $k$-means clustering to obtained feature vectors $\{\bs y_i\}$ is highly expected to improve their performance.  
Obtaining the vector representations $\{\bs y_i\}$ by considering the link weights $\{w_{ij}\}$ is formally called GE.
% and is especially called ``feature learning" if we impose the constraint such as $\bs y_i=f(\bs x_i)$, as we do in this paper.

One simple way to obtain the optimal parameter $\bs {\hat\theta}$ is to minimize the negative log-likelihood
\begin{align}
	L_{0,n}(\bs \theta)
	&:=
	-\log \mathbb{P}(\{w_{ij}\}_{(i,j) \in \mathcal{I}_n} \mid \{\bs x_i\}_{i=1}^{n})
	 \nonumber \\
	&\hspace{-2em}
	\scalebox{0.9}{$\displaystyle 
	=
	\sum_{(i,j) \in \mathcal{I}_n} 
	\bigg\{
	-
	w_{ij} \log \mu_{\bs \theta}(\bs x_i,\bs x_j)
	+
	\mu_{\bs \theta}(\bs x_i,\bs x_j)\bigg\},$}
	\label{eq:likelihood}
\end{align}
which is based on the Poisson model~(\ref{eq:w_model}). 
By minimizing this negative log-likelihood, we obtain the maximum likelihood estimator~(MLE)
\[
\bs{\hat \theta}_0=(\bs{\hat \psi}_0,\hat{\gamma}_0)
:=
\argmin_{\bs \theta \in \bs \Theta} L_{0,n}(\bs \theta),
\]
which specifies the optimal NN $f_{\bs{\hat \psi}_0}$ and the optimal scaling factor $\hat{\gamma}_0 \geq 0$. 
%Stochastic Gradient Descent~(SGD) or minibatch SGD is used for efficiently minimizing the function, and these procedures in detail are shown in \citet{okuno2018probabilistic}. 
The above-mentioned procedure using MLE to obtain feature vectors is called \textit{likelihood-based GE} in this paper.

%In this paper, we especially focus on graph embedding but not general link-weight prediction problems. 
%Therefore, although arbitrary function can be used as the expectation function $\mu(\bs x_i,\bs x_j;\bs \theta)$ appearing in eq.~(\ref{eq:w_model}), we basically consider C-SIPS model~(\ref{eq:csips_model}) in the remaining of this paper. 

\section{$\beta$-GRAPH EMBEDDING}
\label{sec:robust_graph_embedding} 
%%%%%
% 2019/02/08のディスカッションで決まった話の流れ：
% 1. u(q,p)はdensity power-score function, (density power -cross entropy (名前は要確認))
% 2. u(q,p)はこれまで，(kanamoriのscoreも含め)確率モデルやその定数倍程度しか考えてなかった：これをprobability matching beta-score functionと呼ぶ．
% 3. Kawashimaなどはこれを使ってポアソン回帰をやったが計算の困難などが残る
% 4. そこで，確率モデルではなく1st-order moment of $w_{ij}$をmatching する moment-matching \beta-score (u_{\beta}(\mu_*,\mu_{\theta}))を提案する．これはdensity powerそのものだが確率モデルではなくexpectationを考えている点で異なる (1パラグラフ使って強調)
% 5. そのempiricalな表現を提案する． (これはn -> \inftyで収束する)

Although the existing likelihood-based GE explained in Section~\ref{subsec:graph_embedding} had some success, 
MLEs in general are susceptible to contamination in data, because log-likelihood can be strongly influenced by noise. 
%In our setting, link weights can be suffered from incorrect link weights which we call \textit{outlier link weights}.  then MLE can easily be strongly influenced. 
%This influence deteriorates MLE, and it lowers the performance of subsequent statistical analysis, which strongly relies on the nature of feature vectors obtained by graph embedding. 
In Section~\ref{subsec:beta_score}, we review  $\beta$-score and related divergence which have been developed in  robust statistics for robustifying the log-likelihood.
However, naively applying the $\beta$-score to the model~(\ref{eq:w_model}) has two disadvantages as explained in Section~\ref{subsec:beta_div_problems}. 
%in Section~\ref{subsec:problem_presentation}, we formulate the detailed problem set-up. 
To overcome the drawbacks, we introduce \textit{moment $\beta$-score} that robustly measures the difference between the underlying correct expected weights of links and the specified generative model,
and propose its empirical estimation called \textit{empirical moment $\beta$-score~(EMBS)} in Section~\ref{subsec:beta_estimation}. 
Then, we propose robust \textit{$\beta$-GE} equipped with EMBS. 
In Section~\ref{subsec:property}, we theoretically prove that EMBS is robust against noise in link weights. 
In Section~\ref{subsec:minibatch_SGD}, we introduce a minibatch-based efficient stochastic algorithm, and prove that the algorithm locally minimizes EMBS. 
In Section~\ref{subsec:choice_of_beta}, we discuss the selection of $\beta$. 
Finally, we extend $\beta$-GE to a multi-view setting in Section~\ref{subsec:multi_view}. 
%After that, we naturally extend the $\beta$-graph embedding to multi-view setting in Section~\ref{subsec:multi_view}

\textbf{}\subsection{$\beta$-Score for Non-negative Functions}
\label{subsec:beta_score}

Our idea for robustifying the likelihood-based GE is to employ $\beta$-score, that is originally called as density-power score~\citep{kanamori2014affine,kanamori2015robust}.  
% There are, however, two disadvantages when $\beta$-score is naively applied to the probabilistic model~(\ref{eq:w_model}); namely, (i) computationally intractability and (ii) the lack of robustness against distributional misspecification.
% In this section, we describe these two disadvantages in detail, and in Section~\ref{subsec:beta_estimation}, we propose a novel loss function that avoids these disadvantages. 
$\beta$-score robustly measures the difference between two non-negative functions as follows. 

We consider a random variable $\bs z$ taking a value in some set $\mathcal{Z}$,
and consider a set of non-negative functions $\mathcal{P}(\mathcal{Z}):=\{f:\mathcal{Z} \to \mathbb{R}_{\geq 0}\}$.
For non-negative functions $g,f,\nu \in \mathcal{P}(\mathcal{Z})$, we define $\beta$-score as
\begin{align*}
    \scalebox{0.85}{$\displaystyle 
	u_{\beta}^{\bs z}(g,f;\nu)
	:=
	$}
	\begin{cases}
	\scalebox{0.85}{$\displaystyle 
	-\sum_{\bs z \in \mathcal{Z}} g(\bs z)\nu(\bs z)\frac{f(\bs z)^{\beta}-1}{\beta} + \sum_{\bs z \in \mathcal{Z}} \nu(\bs z)\frac{f(\bs z)^{1+\beta}}{1+\beta}$} \\
	\hspace{10em} (\bs z\text{ is discrete}) \\
	\scalebox{0.85}{$\displaystyle 
	-\int_{\mathcal{Z}} g(\bs z)\nu(\bs z)\frac{f(\bs z)^{\beta}-1}{\beta} \mathrm{d}\bs z + \int_{\mathcal{Z}}\nu(\bs z) \frac{f(\bs z)^{1+\beta}}{1+\beta} \mathrm{d}\bs z$}  \\
	\hspace{9em} (\bs z\text{ is continuous}) \\	
	\end{cases}
\end{align*}
where $\beta>0$ is a user-specified tuning parameter. 
For any fixed $g \in \mathcal{P}(\mathcal{Z})$, the minimizer $f \in \mathcal{P}(\mathcal{Z})$ of $u_{\beta}^{\bs z}(g,f;\nu)$ is known to be $f=g$, in the sense that $f(\bs z)=g(\bs z)$ over the support of $\nu$. 
If $\nu(\bs z)=1$ for all $\bs z \in \mathcal{Z}$, 
we abbreviate above $\beta$-score by $u_{\beta}^{\bs z}(g,f)$. 

Let us consider the special case where the functions $f,g$ are restricted to be pdf or pmf, and they are denoted by $q,p$, respectively.
Then, $\beta$-score $u_{\beta}^{\bs z}(q,p)$ is called $\beta$-cross entropy~\citep[see, e.g.][]{futami2018variational}, 
and $D_{\beta}(q,p):=u_{\beta}^{\bs z}(q,p)-u_{\beta}^{\bs z}(q,q)$ is known as $\beta$-divergence~\citep{cichocki2010families} or density-power divergence~\citep{basu1998robust}, which belongs to the Bregman-divergence family~\citep{bregman1967relaxation}. 
In particular, $D_{\beta}(q,p)$ reduces to Kullback-Leibler divergence as $\beta \downarrow 0$. 

$\beta$-score has also been used for unnormalized models~\citep{kanamori2015robust} defined as $c p(\bs z)$ where $c \geq 0$ is a scaling parameter and $p(\bs z)$ is pdf or pmf. For contaminated $q=c_0 q_0 + (1-c_0) r$ with outlier distribution $r$ and contamination rate $1-c_0$, the $\beta$-score $\min_{c\in(0,1]}u_{\beta}^{\bs z}(q,c p)$ is equivalent to the ``$\gamma$-divergence'' between $q$ and $p$, which robustly measures the difference beween $q_0$ and $p$~\citep{jones2001comparison,fujisawa2008robust}.

%$D_{\beta}(q,c \cdot p)$ exactly matches two non-negative functions so that $q = c \cdot p$, 
%$\gamma$-divergence~\citep{fujisawa2008robust} $\tilde{D}_{\gamma}(q,p):=\tilde{d}_{\gamma}(q,p)-\tilde{d}_{\gamma}(q,q)$ defined with $\tilde{d}_{\gamma}(q,p):=h_{\gamma}(\min_{c \geq 0}d_{\gamma}(q,c \cdot p))$, the monotonically increasing function $h_{\gamma}(x):=-\frac{1}{\gamma(1+\gamma)}\log (-\gamma(1+\gamma)x)$, and a user-specified parameter $\gamma>0$, automatically ignores the scaling so that $p$ is proportional to $q$. 

% To distinguish BCE from UBSF, we especially denote BCE by 
% \begin{align*}
% 	d_{\beta}^{\bs z}(q,p)
% 	:=
% 	u_{\beta}^{\bs z}(q,p), 
% \end{align*}
% for two pdfs (or pmfs) $q(\bs z)$ and $p(\bs z)$. 

Although $\beta$-score can measure the difference between arbitrary non-negative functions, it has only been applied to probability distributions and  unnormalized models as seen above.
%\citep{kanamori2014affine,kanamori2015robust,takenouchi2017statistical}. 
We call $\beta$-score considering only these cases as \emph{probability $\beta$-score}.
This is different from our usage of $\beta$-score introduced in Section~\ref{subsec:beta_estimation}.

\textbf{}\subsection{Two Disadvantages of Naively Applying $\beta$-Score to GE}
\label{subsec:beta_div_problems}

Probability $\beta$-score can be employed for GE. 
Considering a conditional pdf $q$ of $w_{ij} \mid \bs x_i,\bs x_j$ and the probabilistic generative model $p_{\bs \theta}$ defined in (\ref{eq:w_model}), we may minimize 
\begin{align}
    E_{\mathcal{X}^2}(u_{\beta}^{(w_{12} \mid \bs x_1,\bs x_2)}(q,p_{\bs \theta})),
    \label{eq:probability_beta_score}
\end{align}
whose empirical estimation is
\begin{align}
&\frac{1}{|\mathcal{I}_n|}
\sum_{(i,j) \in \mathcal{I}_n}
		u_{\beta}^{(w \mid \bs x_i,\bs x_j)}(\hat{q}_{ij},
		p_{\bs \theta}) \nonumber \\
&\hspace{1em}=
-
\frac{1}{|\mathcal{I}_n|}\sum_{(i,j) \in \mathcal{I}_n}
	\frac{p_{\bs \theta}(w_{ij} \mid \bs x_i,\bs x_j)^{\beta}-1}{\beta}\nonumber \\
&\hspace{3em}
+\frac{1}{|\mathcal{I}_n|}\sum_{(i,j) \in \mathcal{I}_n}
	\sum_{w \in \mathbb{N}_0} 
\frac{p_{\bs \theta}(w \mid \bs x_i,\bs x_j)^{1+\beta}}{1+\beta},
\label{eq:empirical_beta_div}
\end{align}
where $\hat{q}_{ij}(w \mid \bs x_i,\bs x_j)$ takes value $1$ if $w=w_{ij}$ and $0$ otherwise. 
(\ref{eq:empirical_beta_div}) is a slight generalization of \citet{ghosh2013robust} that applies probability $\beta$-score to linear regression, and it asymptotically converges in probability to (\ref{eq:probability_beta_score}) as $n \to \infty$. See Lemma~\ref{lem:lln_beta_ce} in Supplement~\ref{appendix:mlln} for the convergence.
% \begin{align}
% &(\ref{eq:empirical_beta_div})
% \overset{p}{\to} 
% E_{\mathcal{X}^2}(u_{\beta}^{(w_{12} \mid \bs x_1,\bs x_2)}(q,p_{\bs \theta})) \quad (n \to \infty), 
% \label{eq:naive_beta_convergence}
% \end{align}
% where $w_{ij} \mid \bs x_i,\bs x_j \overset{\text{indep.}}{\sim} q$ and $E_{\mathcal{X}^2}$ takes expectation over $\{\bs x_i\}$. 
% See Lemma~\ref{lem:lln_beta_ce} in Supplement~\ref{appendix:mlln} for details. 
% Thus minimizing (\ref{eq:empirical_beta_div}) leads to robust estimation of $\bs \theta$ so that the probabilistic model $p_{\bs \theta}$ is empirically matched with the underlying model $q$. 

However, estimating the probabilistic generative model $p_{\bs \theta}$ by minimizing (\ref{eq:empirical_beta_div}) has the following two disadvantages: 
\begin{enumerate}[{(i)}]
\item The last term in (\ref{eq:empirical_beta_div}) is computationally intractable because of the infinite summation $\sum_{w \in \mathbb{N}_0}$.  
\item $w_{ij} \mid \bs x_i,\bs x_j \sim q$ is required to follow Poisson distribution~(\ref{eq:w_model}) for correctly estimating the probabilistic model, but the model is only approximation to the underlying distribution in reality. 
\end{enumerate}

Regarding (i), many of existing studies such as \citet{ghosh2013robust} only consider normal linear regression, so that the corresponding term can be analytically calculated. 
As for non-normal setting, the infinite-summation in (\ref{eq:empirical_beta_div}) similarly appears in eq.~(2.4) of \citet{kawashima2018robust}, that applies $\gamma$-divergence to sparse Poisson regression, and they compute the term by the finite-sum approximation instead.

Regarding (ii), although other probabilistic model can be used as $p_{\bs \theta}$, its estimation is still sensitive to the distributional misspecification as long as the $\beta$-score is naively applied to the user-specified probabilistic generative model $p_{\bs \theta}$.

\subsection{Proposed $\beta$-Graph Embedding}
\label{subsec:beta_estimation}

%To relieve the adverse effect of noisy link weights, UBSF~\citep{kanamori2015robust}-based estimation may be used. 
%However, UBSF using the Poisson-based model~(\ref{eq:w_model}) is hard to optimize, since its bias-correction term requires high computational cost unlike most existing studies that only consider Gaussian-based model. 
%See Supplement~\ref{appendix:ubsf_difficulty} for detail. 
In order to avoid the two disadvantages of probability $\beta$-score explained in Section~\ref{subsec:beta_div_problems}, 
here we introduce
\emph{moment $\beta$-score} that applies $\beta$-score to the expected value of $\bs z$ instead of $p(\bs z)$.
For graph embedding, we consider the conditional expectation of link weights $g(\bs x_1,\bs x_2):=E(w_{12} \mid \bs x_1,\bs x_2)$, and apply $\beta$-sore to it as
\begin{align*}
    &u_{\beta}^{(\bs x_1,\bs x_2)}(g,\mu_{\bs \theta};\nu) = \\
    &\hspace{1em}
    \scalebox{0.95}{$\displaystyle
    E_{\mathcal{X}^2}\left(
    -g(\bs x_1,\bs x_2)\frac{\mu_{\bs \theta}(\bs x_1,\bs x_2)^{\beta}-1}{\beta}
    +
    \frac{\mu_{\bs \theta}(\bs x_1,\bs x_2)^{1+\beta}}{1+\beta}
    \right),$}
\end{align*}
where $\nu$ is pdf of $(\bs x_1,\bs x_2) \in \mathcal{X}^2$. 
$\beta>0$ is a user-specified tuning parameter, and 
$\mu_{\bs \theta}$ is specified as C-SIPS~(\ref{eq:csips_model}).

Moment $\beta$-score can be empirically estimated by 
\emph{empirical moment $\beta$-score~(EMBS)}, that is
\begin{align}
&L_{\beta,n}(\bs \theta):= \nonumber\\
&\hspace{0.5em}
\scalebox{0.9}{$\displaystyle 
\sum_{(i,j) \in \mathcal{I}_n} 
\bigg\{
	-
	w_{ij} \frac{\mu_{\bs \theta}(\bs x_i,\bs x_j)^{\beta}-1}{\beta}
+
\frac{
	\mu_{\bs \theta}(\bs x_i,\bs x_j)^{1+\beta}
}{1+\beta}
\bigg\}$}
\label{eq:beta_loss}
\end{align}
in the sense that 
$L_{\beta,n}(\bs \theta)
	\overset{p}{\to}
	u_{\beta}^{(\bs x_1,\bs x_2)}(g,\mu_{\bs \theta};\nu)$ as $n \to \infty$, 
% 	\label{eq:proposed_beta_convergence} 
% \end{align}
under some assumptions. 
See Lemma~\ref{lem:lln_EMBS} in Supplement~\ref{appendix:mlln} for the convergence.

Noisy link weights can be modeled by defining the conditional expectation $g(\bs x_i,\bs x_j)=E(w_{ij} \mid \bs x_i,\bs x_j)$ as the sum of underlying  weight $\mu_*(\bs x_i,\bs x_j)$ and noise  $\eta_*(\bs x_i,\bs x_j)$. 
As moment $\beta$-score robustly measures the difference between $\mu_*$ and $\mu_{\bs \theta}$ even if $g$ is contaminated by $\eta_*$, EMBS is robust against noise in link weights. See Section~\ref{subsec:property} for details.

In addition to the robustness of EMBS against noisy link weights, EMBS simultaneously avoids the two disadvantages of probability $\beta$-score explained in Section~\ref{subsec:beta_div_problems}. 
Regarding (i), i.e.~the computational intractability,  the numerical optimization of $L_{\beta,n}(\bs \theta)$ is not difficult since infinite summation is not involved.
Regarding (ii), i.e.~the lack of robustness against distributional misspecification, EMBS is free from this problem because it does not require distributional specification.

%EMBS asymptotically measures the moment $\beta$-score between only the expected weight functions $g$ and $\mu_{\bs \theta}$, unlike (\ref{eq:probability_beta_score}) that measures probability $\beta$-score between two pmfs $q$ and $p_{\bs \theta}$. 

%(\ref{eq:proposed_beta_convergence}) is obtained by the law of large numbers for doubly-indexed partially dependent random variables, that is shown in Supplement~\ref{appendix:mlln}. 

We replace the negative log-likelihood $L_{0,n}(\bs \theta)$ defined in (\ref{eq:likelihood}) with EMBS $L_{\beta,n}(\bs \theta)$ defined in (\ref{eq:beta_loss}). 
The remaining procedure is all the same: 
we define the \textit{$\beta$-estimator} as a minimizer of EMBS
\begin{align}
\bs{\hat \theta}_{\beta}
=
(\bs{\hat \psi}_{\beta},\hat{\gamma}_{\beta})
:=
\argmin_{\bs \theta \in \bs \Theta}
L_{\beta,n}(\bs \theta),
\label{eq:beta_estimator}
\end{align}
and the $\beta$-estimator defines feature vectors $\bs y_{\beta,i}:=f_{\bs{\hat \psi}_{\beta}}(\bs x_i)$, $i=1,2,\ldots,n$ by substituting $\bs{\hat \psi}_{\beta}$ into (\ref{eq:transformation}). 
%\begin{align}
%\mathbb{R}^p
%\supset
%\mathcal{X}
%\ni
%\bs x_i
%\mapsto
%\bs y_{\beta,i}
%:=
%f_{\bs{\hat \psi}_{\beta}}(\bs x_i)
%\in
%\mathbb{R}^K.
%\label{eq:beta_embedding}
%\end{align}
The whole procedure described above, that obtains feature vectors using the $\beta$-estimator (\ref{eq:beta_estimator}), is called \textit{$\beta$-graph embedding~($\beta$-GE)}. 
Since EMBS reduces to the negative log-likelihood as $\beta \downarrow 0$, $\beta$-GE naturally extends the likelihood-based GE, and its robustness can be formally proven as shown in the following section.

\subsection{Robustness against Noise in Link Weights}
\label{subsec:property}

Thus far, we have described likelihood-based GE using the model~(\ref{eq:w_model}) and proposed $\beta$-GE. 
However, the model~(\ref{eq:w_model}) does not explain how noisy link weights are actually produced. 
%Here, we here study the nature of $\beta$-estimator. 
For that reason, in this section, we first formally define the probabilistic model that considers noisy link weights. 
Then, we explain why moment $\beta$-score, that is used in $\beta$-GE, is robust against noise in link weights. 
The following explanation for moment $\beta$-score is an adaptation of that for probability $\gamma$-score given in Lemma 3.1 of \citet{fujisawa2008robust}.

%Noisy link weights can be formulated in a similar way of \citet{fujisawa2008robust} and \citet{kanamori2014affine}. 
We consider the generative model of $w_{ij}\ge0$ up to the first and second moments.
For describing noisy link weights, 
\textit{expected noise} $\eta_*(\bs x_1,\bs x_2)$ is added to the underlying \textit{correct expected weight} $\mu_*(\bs x_1,\bs x_2)$, where
$\mu_*$ and $\eta_*$ are non-negative functions over $\mathcal{X}^2$.
%% rateではなくcoefficient (mixtureになってるわけではなく，単に足してるだけだから・・)
\begin{align}
E(w_{ij} \mid \bs x_i,\bs x_j)
&=
\mu_*(\bs x_i,\bs x_j)
+
\eta_*(\bs x_i,\bs x_j)
\nonumber \\
&=:
g(\bs x_i,\bs x_j),  \label{eq:w_cont_model} \\
E(w_{ij}^2 \mid \bs x_i,\bs x_j)
&<\infty,
\label{eq:w_var} \\
\bs x_i &\overset{\text{i.i.d.}}{\sim} Q
\label{eq:x_model} 
\end{align}
for all $(i,j) \in \mathcal{I}_n$. 
The support of the density function $Q(\bs x)$ is denoted as $\text{supp}Q \subset \mathcal{X}$.
The amount of noise is measured by
\begin{align} 
\alpha:=E_{\mathcal{X}^2}(\eta_*(\bs x_1,\bs x_2)),
\label{eq:scale_of_eta_and_mu}
\end{align}
%$\mu_*,\eta_*$ take positive values for all $(\bs x_1,\bs x_2) \in (\text{supp}Q)^2$, 
%$\mu_*$ and $\eta_*$ are normalized so that $E_{\mathcal{X}^2}(\eta_*(\bs x_1,\bs x_2))=E_{\mathcal{X}^2}(\mu_*(\bs x_1,\bs x_2))=O(1),E_{\mathcal{X}^2}(\eta_*(\bs x_1,\bs x_2)^2)<\infty,E_{\mathcal{X}^2}(\mu_*(\bs x_1,\bs x_2)^2)<\infty$ 
where $E_{\mathcal{X}^2}$ indicates expectation with respect to the joint density $\nu(\bs x_1, \bs x_2) := Q(\bs x_1) Q(\bs x_2)$.
% The correct link weights are specified by $\mu_*$, and they are disturbed by $\eta_*$.
%% This type of mis-labelling (of link weights in our setting) is classified into noisy not at random model defined in \citet{frenay2014classification}. 
%% mislabelling mechanismって単語あるのか?
%% 一応精査する -> 証明もappendixに一応載せる?
%The model~(\ref{eq:w_cont_model}) is easily interpretable:
%Regardless of whether the true expectation function $\mu_*(\bs x_i,\bs x_j)$ is large or not, $w_{ij}$ is likely to be positive if the outlier expectation function $\eta_*(\bs x_i,\bs x_j)$ is large. 
%%Thus we call $\eta_*$ as outlier expectation function. 
%%and we describe this contamination as ``link weights are contaminated by outlier link weights" in the remaining of this paper.
We implicitly assume that $\mu_*(\bs x_1, \bs x_2)$ is sufficiently small when $\eta_*(\bs x_1, \bs x_2)$ is large. 
More specifically, supposing 
\begin{align}
	E_{\mathcal{X}^2}(\eta_*(\bs x_1,\bs x_2)\mu_*(\bs x_1,\bs x_2)^{\beta_0})
	<
	\varepsilon_*,
	\label{eq:characterization_outlier}
\end{align}
we assume that $\varepsilon_*$ is sufficiently small for an appropriately large $\beta_0>0$. This corresponds to the assumption ($\ast$) for $\nu_f$ in \citet{fujisawa2008robust}.
We also assume that the model $\mu_{\bs \theta}(\bs x_i,\bs x_j)$ is correctly specified:
\begin{align}
\exists \bs \theta_* \in \bs \Theta \: \text{such that} \:
\mu_{\bs \theta_*}(\bs x_1,\bs x_2)
=
\mu_*(\bs x_1,\bs x_2)
\label{eq:theta_star}
\end{align}
for all $(\bs x_1,\bs x_2) \in (\text{supp}Q)^2$.

Following the above settings, $\beta$-GE robustly infers the correct expected weight $\mu_*$ from noisy link weights $\{w_{ij}\}$ whose conditional expectation is contaminated by noise as (\ref{eq:w_cont_model}).

%Similarly to the robustness of EMBS, $\beta$-estimator has following four advantages: 

%\begin{defi}[Robustness against outliers]
%aa
%\end{defi}
%
%\begin{defi} Expectation function $\mu(\bs x_1,\bs x_2;\bs{\hat \theta})$ estimated with $n$ samples is said to be \textit{expectation consistent} if $\mu(\bs x_1,\bs x_2;\bs{\hat \theta}) \overset{p}{\to} \mu_*(\bs x_1,\bs x_2)$ defined in (\ref{eq:w_cont_model}) with $\alpha=0$. 
%\end{defi}
%
%
%\begin{defi}[Distribution-free]
%If the estimation of expectation function $\mu(\bs x_1,\bs x_2;\bs{\hat \theta})$ is distribution-free if it is expectation-consistent without assuming any distribution on $\{w_{ij}\}$ other than the designation of its first- and second-order moment.  
%\end{defi}

%(i) \textbf{robust} against outliers, (ii) \textbf{distribution-free}, (iii) \textbf{expectation consistent} regardless of the value $\beta>0$, and (iv) naturally \textbf{extends MLE}.
%We first present these properties (i)--(iii) in following Theorem~\ref{theo:robust_consistency}. 
%After that, we show that $\beta$-estimator is a natural extension of MLE in Proposition~\ref{prop:beta_est_to_mle}. 

For identifying the robustness of $\beta$-GE, we consider a restricted parameter set 
\begin{align}
&\bs \Theta_{\varepsilon}:= 
\{\bs \theta \in \bs \Theta
\mid \nonumber \\
&\hspace{5em}
E_{\mathcal{X}^2}
\left(
	\eta_*(\bs x_1,\bs x_2)\mu_{\bs \theta}(\bs x_1,\bs x_2)^{\beta_0}
\right)
<
\varepsilon
\} 
\label{eq:robust_parameter_set}
\end{align}
that satisfies $\bs \theta_* \in \bs \Theta_{\varepsilon_*}
\subseteq \bs \Theta_{\varepsilon}$ for  $\varepsilon \geq \varepsilon_*$
if (\ref{eq:characterization_outlier}) and (\ref{eq:theta_star}) hold.

%According to the law of large numbers for doubly-indexed partially-dependent random variables shown in Supplement~\ref{appendix:mlln}, the EMBS converges
%\begin{align}
%&
%|\mathcal{I}_n|^{-1}L_{\beta,n}(\bs \theta) \nonumber \\
%&\hspace{1em} \overset{p}{\to} 
%d_{\beta}(E(w_{12} \mid \bs x_1,\bs x_2),\mu(\bs x_1,\bs x_2;\bs \theta)) \nonumber \\
%&\hspace{1em}
%%d_{\beta}(E(w_{12} \mid \bs x_1,\bs x_2),\mu(\bs x_1,\bs x_2;\bs \theta);q(\bs x_1)q(\bs x_2))
%:=E_{\mathcal{X}^2}\bigg(
%	-\frac{1}{\beta} E(w_{12} \mid \bs x_1,\bs x_2) \left( \mu(\bs x_1,\bs x_2;\bs \theta)^{\beta}-1 \right) \nonumber \\
%	&\hspace{8em}
%	+
%	\frac{1}{1+\beta} \mu(\bs x_1,\bs x_2;\bs \theta)^{1+\beta}
%\bigg)
%\label{eq:asymptotic_lbeta}
%\end{align}
%as $n\to\infty$, where $E_{\mathcal{X}^2}$ represents to take expectation over $(\bs x_1,\bs x_2)$. 
%The function $d_{\beta}(\cdot,\cdot)$ is known as UBSF~\citep{kanamori2015robust}, which is a robust discrepancy measure between two non-negative functions. 
% %When considering the mixture of ``true" non-negative function $f$ and the ``abnormal" non-negative function $\xi$ which is far different from $f$, $d_{\beta}(f+\alpha \xi,g_{\bs \theta}) \approx d_{\beta}(f,g_{\bs \theta})$ holds for some $\bs \theta,\alpha \geq 0$ and some functions $g_{\bs \theta}$. 

\begin{theo}
\label{theo:div_perturbation}
Suppose that $\varepsilon \geq \varepsilon_*,\beta \in (0,\beta_0]$, and (\ref{eq:w_cont_model})--(\ref{eq:robust_parameter_set}) hold. 
Then, there exists a function $M(\bs \theta) \geq 0$ such that, for all $\bs \theta \in \bs \Theta_{\varepsilon}$,
\begin{align}
%u_{\beta}^{(\bs x_1,\bs x_2)}(g,\mu_{\bs \theta}) 
L_{\beta,n}(\bs \theta)
&=
u_{\beta}^{(\bs x_1,\bs x_2)}(\mu_*,\mu_{\bs \theta};\nu)
+
\alpha \beta^{-1}
\nonumber \\
&\hspace{2.5em}
-
M(\bs \theta) \, \varepsilon^{\beta/\beta_0}
+
O_p(1/\sqrt{n}),
\label{eq:theo_div_perturbation} \\
M(\bs \theta) &\leq \alpha^{1-\beta/\beta_0}\beta^{-1}.
\label{eq:M_beta_evaluation}
\end{align}
\end{theo}

\textbf{Proof:} The law of large numbers shown in Lemma~\ref{lem:lln_EMBS} of Supplement indicates $L_{\beta,n}(\bs \theta) = u_{\beta}^{(\bs x_1,\bs x_2)}(g,\mu_{\bs \theta}; \nu)+O_p(1/\sqrt{n})$, and a simple calculation leads to (\ref{eq:theo_div_perturbation}) as
\begin{align*}
&
\scalebox{0.9}{$\displaystyle 
u_{\beta}^{(\bs x_1,\bs x_2)}(g,\mu_{\bs \theta};\nu) 
=
    u_{\beta}^{(\bs x_1,\bs x_2)}(\mu_* + \eta_*,\mu_{\bs \theta};\nu)$} \\
&\hspace{1em}
\scalebox{0.9}{$\displaystyle 
=
u_{\beta}^{(\bs x_1,\bs x_2)}(\mu_*,\mu_{\bs \theta};\nu) 
-
E_{\mathcal{X}^2}\Bigl(
\eta_*(\bs x_1,\bs x_2)
    \frac{\mu_{\bs \theta}(\bs x_1,\bs x_2)^{\beta}-1}{\beta}
\Bigr)$} \\
&\hspace{1em}
\scalebox{0.9}{$\displaystyle 
=
u_{\beta}^{(\bs x_1,\bs x_2)}(\mu_*,\mu_{\bs \theta};\nu) 
+
\underbrace{
\beta^{-1}
E_{\mathcal{X}^2}\left(
	\eta_*(\bs x_1,\bs x_2)
\right)}_{=\alpha \beta^{-1}}$} \\
&\hspace{2em}
\scalebox{0.9}{$\displaystyle 
-
\underbrace{\beta^{-1} E_{\mathcal{X}^2}(\eta_*(\bs x_1,\bs x_2)\mu_{\bs \theta}(\bs x_1,\bs x_2)^{\beta})\varepsilon^{-\beta/\beta_0}}_{=:M(\bs \theta)}
\varepsilon^{\beta/\beta_0}.$}
\end{align*}
Then (\ref{eq:M_beta_evaluation}) follows from the inequality in (\ref{eq:robust_parameter_set}) and Lyapunov's inequality as shown in Lemma~\ref{lem:order_epsilon} of Supplement. \qed

Theorem~\ref{theo:div_perturbation} asserts that, EMBS $L_{\beta,n}(\bs \theta)$ approximates 
$u_{\beta}^{(\bs x_1,\bs x_2)}(\mu_*,\mu_{\bs \theta};\nu)$ up to the constant $\alpha \beta^{-1}$ as long as $\varepsilon>0$ is specified to be sufficiently small, 
even if the correct expected weight $\mu_*$ is contaminated by expected noise $\eta_*$. 
% The term $M(\bs \theta) \cdot \varepsilon^{\beta/\beta_0}$ can be ignored. 
As $\alpha \beta^{-1}$ is constant with respect to $\bs \theta$, and $u_{\beta}^{(\bs x_1,\bs x_2)}(\mu_*,\mu_{\bs \theta})$ is minimized at $\bs \theta=\bs \theta_*$, minimization of EMBS leads to robust estimation of the correct expected weight  $\mu_*$.

\subsection{Minibatch-Based Stochastic Optimization Algorithm}
\label{subsec:minibatch_SGD}
To minimize the EMBS~(\ref{eq:beta_loss}), we consider gradient descent-based approaches. 
However, computing a full-batch gradient in our setting requires summing up $O(|\mathcal{I}_n|)=O(n^2)$ terms for each iteration. 
In practice, the computational complexity is remarkably high and non-negligible. 
To reduce this high computational complexity, we employ a minibatch-based stochastic algorithm.

Let $\mathcal{W}_n$ be an index set of positive weights $\{(i,j) \in \mathcal{I}_n \mid w_{ij}>0\} \subset \mathcal{I}_n$. 
At iteration $t=1,2,\ldots$, we pick up $m_1,m_2 \in \mathbb{N}$ elements from $\mathcal{W}_n,\mathcal{I}_n$ uniformly at random, and denote the sets as 
$\mathcal{W}_n^{(t)},\mathcal{I}_n^{(t)}$, respectively. Here 
$\mathcal{W}_n^{(t)}$ and $\mathcal{I}_n^{(t)}$ can overlap, 
but no duplication in each set. 
In the remaining of this section, $E^{(t)}, E^*$ represent expectations with respect to resampling sets $(\mathcal{W}_n^{(t)},\mathcal{I}_n^{(t)}), 
\{(\mathcal{W}_n^{(t)},\mathcal{I}_n^{(t)})\}_{t=1,2,\ldots}$, respectively.

Similarly to \citet{okuno2018probabilistic} Section~4, EMBS is stochastically approximated by 
\begin{align*}
h^{(t)}(\bs \theta)
&:=
%  \frac{\partial}{\partial \bs \theta}
%  \bigg[
-
\sum_{(i,j) \in \mathcal{W}_n^{(t)}} 
w_{ij} 
\frac{\mu_{\bs \theta}(\bs x_i,\bs x_j)^{\beta}-1}{\beta}  \\
&\hspace{5em}
+
\lambda
\sum_{(i,j) \in \mathcal{I}_n^{(t)}} 
\frac{\mu_{\bs \theta}(\bs x_i,\bs x_j)^{1+\beta}}{1+\beta}, 
\end{align*}
where $\lambda > 0$ is a tuning parameter. 
Then, our iterative algorithm updates $\bs \theta$ by
\begin{align}
\bs \theta^{(t+1)}:=
\mathcal{P}_{\bs \Theta}\left( 
    \bs \theta^{(t)} - \delta^{(t)} \frac{\partial h^{(t)}(\bs \theta)}{\partial \bs \theta} \bigg|_{\bs \theta=\bs \theta^{(t)}}
\right)
\label{eq:minibatch_sgd_formulation}
\end{align}
with learning rate $\delta^{(t)} \geq 0$, $t \in \mathbb{N}_0$, user-specified initial value $\bs \theta^{(0)}$, and projection function $\mathcal{P}_{\bs \Theta}(\bs \theta'):=\argmin_{\bs \theta \in \bs \Theta}\|\bs \theta-\bs \theta'\|_2^2$. 
This algorithm is a slight modification of minibatch stochastic gradient descent~\citep[minibatch SGD; see][]{goodfellow2016deep} with a flavor of negative sampling~\citep{mikolov2013distributed}. 
A similar algorithm for likelihood-based GE can be found in~\citet{okuno2018probabilistic}. 
%Minibatch SGD requires only a constant number of operations for computing gradient in each step; minibatch SGD much reduces the computational complexity. 

Compared with the plain SGD that uses only one sample for computing the gradient in each step, minibatch-based stochastic algorithm is proved to be more stable, in the sense that its asymptotic variance with respect to the number of iterations is smaller~\citep{bonakdarpour2016statistical,toulis2017asymptotic}. 
Although our proposed algorithm~(\ref{eq:minibatch_sgd_formulation}) involves the randomness to make sets $\mathcal{W}_n^{(t)},\mathcal{I}_n^{(t)} \: (t=1,2,\ldots)$, its convergence limit can be established as shown in the following theorem.

\begin{theo}
\label{theo:minibatchSGD_limit} 
In order to consider the optimization locally, we  redefine the parameter set as $\bs \Theta = \{\bs \theta \in \mathbb{R}^q \mid \|\bs \theta - \bs\theta_0 \|_2 \le D\}$ with a constant $D>0$ and a fixed parameter value $ \bs\theta_0 \in \mathbb{R}^q$.
Suppose that $\bs \theta_* \in \bs\Theta$ is a solution of  $\partial h(\bs\theta)/\partial \bs \theta = \bs 0$ for $ h(\bs \theta):=E^{(1)}(h^{(1)}(\bs \theta))$.
We also assume that
(i) $\mu_{\bs \theta}(\bs x_1,\bs x_2) \in C^1(\bs \Theta)$ for any $(\bs x_1,\bs x_2) \in \mathcal{X}$, 
(ii) $h(\bs \theta)$ is strongly-convex on $\bs \Theta$, and 
(iii) $\delta^{(t)}=O(t^{-\alpha})$ for some $\alpha \in (0,1]$. 
Then, the estimator $\bs \theta^{(t)}$ of our proposed algorithm~(\ref{eq:minibatch_sgd_formulation}) converges to $\bs \theta_*$, in the sense that $E^*(\|\bs \theta^{(t)}-\bs \theta_*\|_2^2) \to 0$ as $t \to \infty$.
The solution $\bs\theta_*$ is the unique minimizer of $h(\bs\theta)$ over $\bs\Theta$, and the equation
$\partial h(\bs\theta)/\partial \bs \theta = \bs 0$ 
is written as
 \begin{align}
 &
 \scalebox{0.9}{$\displaystyle 
 \sum_{(i,j) \in \mathcal{I}_n}
 \bigg\{
 \bigg(
 -
 v m_1 w_{ij} 
 +
 \lambda m_2 \mu_{\bs \theta}(\bs x_i,\bs x_j)
 \bigg)$} \nonumber \\
 &\hspace{4em}\times 
 \scalebox{0.9}{$\displaystyle
 \mu_{\bs \theta}(\bs x_i,\bs x_j)^{\beta}
 \frac{\partial \log \mu_{\bs \theta}(\bs x_i,\bs x_j)}{\partial \bs \theta}
 \bigg\} = \bs 0,$}
 \label{eq:minibatch_estimating}
 \end{align} 
where $v:=|\mathcal{I}_n|/|\mathcal{W}_n|$, $m_1:=|\mathcal{W}_n^{(1)}|$, and $m_2:=|\mathcal{I}_n^{(1)}|$. 
\end{theo}

As (\ref{eq:minibatch_sgd_formulation}) is classified as a standard projected stochastic gradient descent, 
% which has been well studied for more than half a century, 
applying existing theorems such as \citet{eric2011non} Theorem~2 leads to Theorem~\ref{theo:minibatchSGD_limit}. 
% \citet{benveniste1990adaptive} Theorem 22 leads to Theorem~\ref{theo:minibatchSGD_limit}. 
% See, for example, \citet{kushner2003stochastic} Section 1.1.1 which asserts that, as $t \to \infty$,  (\ref{eq:minibatch_sgd_formulation}) converges in probability to the solution of $E^{(1)}(\bs H^{(1)}(\bs \theta))=\bs 0$ where $E^{(1)}$ represents expectation over the randomness to make sets $\mathcal{W}_n^{(1)},\mathcal{I}_n^{(1)}$. 
Proof and further explanation are given in Supplement~\ref{proof_minibatchSGD}.

Theorem~\ref{theo:minibatchSGD_limit} indicates that the convergence limit of our proposed algorithm satisfies the estimating equation $\partial L_{\beta,n}(\bs \theta)/\partial \bs \theta=\bs 0$ if $\lambda m_2=vm_1$. Then it locally minimizes $L_{\beta,n}(\bs \theta)$.

Moreover, even if $\lambda m_2 \neq v m_1$,  (\ref{eq:minibatch_estimating}) indicates that $w_{ij}$ is approximated by $k\,  \mu_{\bs \theta}(\bs x_i,\bs x_j)$ with $k=\lambda m_2/vm_1$,
because (\ref{eq:minibatch_estimating}) is equivalent to replacing $\mu_{\bs \theta}$ by $k\mu_{\bs \theta}$ in (\ref{eq:beta_loss}).
Then $\mu_{\bs \theta}(\bs x_i,\bs x_j)$ approximates the link weight $w_{ij}$ up to the scaling $k$.
In practice, the scaling is not really an issue; we only need the ratio $\mu_{\bs \theta}(\bs x_i,\bs x_j)/\mu_{\bs \theta}(\bs x_k,\bs x_l)$ to infer which of the pairs $(\bs x_i,\bs x_j),(\bs x_k,\bs x_l)$ has a stronger relation. 
Thus, we can ignore the condition $\lambda m_2=vm_1$, and we empirically determine $\lambda$ in experiments in order to seek faster convergence of the algorithm.

\subsection{Selection of the Parameter $\beta$} 
\label{subsec:choice_of_beta}
In this section, we discuss the selection of $\beta>0$. 
As Theorem~\ref{theo:div_perturbation} shows, 
the bias term in (\ref{eq:theo_div_perturbation}) is $O(\varepsilon^{\beta/\beta_0})$ if $\bs \theta \in \bs \Theta_{\varepsilon}$. 
Thus larger $\beta$ may make the term smaller; a larger $\beta$ enhances the robustness. 
On the other hand, as $\beta \downarrow 0$, the $\beta$-estimator converges to the MLE, which is known to be asymptotically efficient~\citep{ferguson1996course}; a smaller $\beta$ enhances the asymptotic efficiency. 
To simultaneously attain high robustness and high asymptotic efficiency, $\beta$ needs to be specified properly. 
In fact, specifying a proper $\beta$ has been a central issue when considering UBSF-related methods~\citep{durio2011minimum,ghosh2015robust} for the same reason.  
However, no decisive way has been presented, even when considering simple linear regression.

Referring to existing studies on a tuning parameter selection of the $\beta$-score, an idea worth considering is applying cross-validation~(CV) with EMBS using a fixed parameter $\beta_0>0$ (independent of $\beta$).
%but not with the negative log-likelihood. 
%EMBS-based $K$-fold CV is defined as 
%\begin{align}
%&\text{CV}(\beta;\beta_0) 
%:=
%\frac{1}{K}
%\sum_{k=1}^{K}
%L_{\beta_0,k}(\bs{\hat \theta}_{\beta,k}), 
%\label{eq:beta_cv}
%\end{align}
%where $\beta_0>0$ is a user-specified constant independent of $\beta$, 
%$\mathcal{I}_{n,k}$ is a $k$-th partition of $\mathcal{I}_{n}$, $L_{\beta_0,k}(\bs \theta)$ is EMBS using $\beta_0$ defined over $\mathcal{I}_{n,k}$ instead of $\mathcal{I}_n$, and $\bs{\hat \theta}_{\beta,k}$ is the $\beta$-estimator trained with $\{w_{ij}\}_{(i,j) \in \mathcal{I}_n \setminus \mathcal{I}_{n,k}}$ $(k=1,2,\dots,K)$. 
By virtue of the robustness of EMBS, 
this EMBS-based CV 
%(\ref{eq:beta_cv}) 
is expected to be robust against noisy link weights whereas a negative log-likelihood-based CV may not be. 
A similar idea can be found in \citet{mollah2007robust} and \citet{kawashima2017robust}.

However, CV requires significant computational resources, and existing studies on the UBSF in fact empirically demonstrate that the UBSF with $\beta$ heuristically chosen from $[0.1,1.0]$ even demonstrates a good performance in practice~\citep{jian2005robust,dessein2010real}. 
%Therefore, in this paper, we heuristically choose $\beta$ from $\{0.1,0.5,1.0\}$ without conducting CV in experiments. 

%%%% uniquenssなくても大丈夫の話
%%Although traditional statistics paid much attention to the uniqueness of estimators, 
%%In this paper, we consider 
%Regarding the uniqueness of estimators, we last discuss to note the non-uniqueness of $\beta$-estimators in our model. 
%Since C-SIPS model~(\ref{eq:csips_model}) is generally having many redundant parameters, the minimizer of $\beta$-loss function with C-SIPS is in general not unique. 
%However, turning our eyes into the estimation of the ``expectation function" $\mu(\bs x_i,\bs x_j;\bs \theta)$, 
%its approximation bound is obtained by Theorem~\ref{theo:robust_consistency} without assuming the uniqueness of $\beta$-estimators. 
%The estimated expectation function $\mu(\bs x_i,\bs x_j;\bs{\hat \theta}_{\beta})$ approximates the true expectation function $\mu_*(\bs x_i,\bs x_j)$ up to scaling constant $(1-\alpha)$ regardless of whether $\beta$-estimator (\ref{eq:beta_estimator}) is unique or not. 
%Thus, we do not especially care the uniqueness throughout this paper. 

\subsection{Extension to the Multi-View Setting}
\label{subsec:multi_view}

Here, we consider the \textit{multi-view} setting, that there exist $D$ different types of data vectors, 
whereas the above-mentioned methods consider only one type. 
We denote the type of $i$th vector $\bs x_i$ as $d_i \in \{1,2,\ldots,D\}$, and the dimension $p_{d_i}$ of $\bs x_i$ depends on the type $d_i$. 
A typical example of multi-view data is images and their text tags, whose associations are represented by links.

To employ such multi-view data vectors, (\ref{eq:csips_model}) can be simply extended to the multi-view setting as
\begin{align}
&\log \mu_{\bs \theta}(\bs x_i,\bs x_j) \nonumber \\
&\hspace{3em}
=
	\langle 
		f_{\bs \psi^{(d_i)}}^{(d_i)}(\bs x_i)
		,
		f_{\bs \psi^{(d_j)}}^{(d_j)}(\bs x_j)
	\rangle
	-
	\gamma^{(d_i,d_j)},
	\label{eq:c_sips_multiview}
\end{align}
where $f^{(d)}_{\bs \psi^{(d)}}:\mathbb{R}^{p_d} \to \mathbb{R}^{K}, \: (d=1,2,\ldots,D)$ are different NNs whose parameters are $\{\bs \psi^{(d)}\}_{d=1}^{D}$, and $K$ is specified by users such that $K \leq p_1,p_2,\ldots,p_d$. 
As we also consider the scaling parameter $\{\gamma^{(d,e)}\}_{1 \leq d < e \leq D}$, the full parameter vector is $\bs \theta=(\{\bs \psi^{(d)}\}_{d=1}^{D},\{\gamma^{(d,e)}\}_{1 \leq d < e \leq D})$. 
This multi-view extension of likelihood-based GE using (\ref{eq:c_sips_multiview}) is proposed recently and it is called probabilistic multi-view GE~\citep[PMvGE; see][]{okuno2018probabilistic}. 

By specifying linear transformations for $f^{(d)}_{\bs \psi^{(d)}}$, PMvGE  approximates CDMCA~\citep{shimodaira2016cross} that generalizes various multivariate analysis methods such as principal component analysis (PCA), canonical correlation analysis~\citep[CCA; see][]{hotelling1936relations}, LPP, HIMFAC~\citep{nori2012multinomial}; see Section 3.6 of \citet{okuno2018probabilistic}. 
Thus, PMvGE extends various multivariate analysis methods. 
%In addition, Poisson regression~\citep{frome1973regression} is obviously a special case of PMvGE as shown in Supplement~\ref{subsec:poisson_regression}. 

The multi-view setting of PMvGE is easily incorporated into $\beta$-GE by (\ref{eq:c_sips_multiview}).
This multi-view $\beta$-GE corresponds to robustification of PMvGE, and therefore it indeed robustifies various existing methods for multi-view analysis.

\section{EXPERIMENTS}
\label{sec:experiments}

In order to confirm the robustness of $\beta$-GE, we conduct numerical experiments on a synthetic dataset in Section~\ref{subsec:numeraical_experiment} and a real-world dataset in Section~\ref{subsec:real_experiment_1}.
Feature vectors obtained by graph embedding methods are evaluated by clustering task.
% and \ref{subsec:real_experiment_2}. 

\subsection{Experiment on a Synthetic Dataset}
\label{subsec:numeraical_experiment}

\textbf{Synthetic dataset:} 
For generating $n=200$ data vectors $\{\bs x_i\}_{i=1}^{200}$ of $p=20$ dimensions, we first prepared centers of clusters $\bs x^{(0)}_k \in \mathbb{R}^5$,  $k=1,\ldots,4$, and a linear transformation $\bs A \in \mathbb{R}^{20 \times 5}$, where all the elements are generated from $N(0,1)$ independently.
Then data vectors are generated by $\bs x_i \overset{\text{i.i.d.}}{\sim} N(\bs A \bs x^{(0)}_{k_i},\bs I_{20})$ for $i=1,2,\ldots,200$, where $k_i:=\lceil i/50 \rceil$ represents cluster index to which $\bs x_i$ belongs.
$\{\bs x_i\}_{i=1}^{50}$ are finally rescaled such that $\sum_{i=1}^{200} \|\bs x_i\|_2/200=4$. Then binary link weights $\{w_{ij}\}_{i,j=1}^{200}$ are generated from Bernoulli distribution. For $k_i=k_j$, $w_{ij}=w_{ji} \sim B(0.05)$ ($i\neq j$) and $w_{ii}=0$.
For $k_i\neq k_j$, $w_{ij}=w_{ji}\sim B(\xi)$ with a specified parameter $\xi \in [0,1]$.

\textbf{Estimation:} 
Feature vectors $\{\bs y_i\}$ are computed by applying likelihood-based GE (corresponding to $\beta=0$) and $\beta$-GE ($\beta=0.1,0.5,1$) to the generated data vectors $\{\bs x_i\}$ and link weights $\{w_{ij}\}$. 
Loss functions are ridge-regularized.
For optimization, we utilize the BFGS algorithm~\citep{fletcher2013practical} with random initialization.

\textbf{Evaluation:} 
We apply $k$-means clustering to obtained feature vectors, and evaluate the result using the purity score~\citep{manning2008introduction}. 

%Using synthetic dataset, 
%$2$-dim. feature vectors of nodes, that are linearly transformed from observed $20$-dim. data vector representations, are plotted with their observed links. 
%Nodes are mainly linked within each of separately-coloured 4 clusters, but there are some links across clusters. 
%We regard these links across clusters as \textit{noisy links}. 
%In each figure \subref{fig:likelihood}--\subref{fig:beta1}, linear transformation is computed by likelihood-based GE and $\beta$-GE with random initialization. 
%BFGS algorithm~\citep{fletcher2013practical} with random initialization is used for optimization. 

\textbf{Results:} 
Sample average and standard deviation of purity scores over 10 experiments are listed in Table~\ref{table:purity} along with noisy link probability $\xi$. 
The purity score takes a value in $[0,1]$ and a higher value is better. 
When $\xi$ is small $(\xi=0.01)$, 
likelihood-based GE achieves the best score, however, if the probability increases ($\xi=0.02,0.03$), 
$\beta$-GE with larger $\beta$ achieves a better score than likelihood-based GE.

\begin{table}[htbp]
\centering
\caption{Sample average and standard error of purity scores over 10 experiments. Higher score is better.}
\label{table:purity}
\scalebox{0.9}{
\begin{tabular}{lccc}
\hline
 & $\xi=0.01$ & $\xi=0.02$ & $\xi=0.03$ \\
\hline
Likelihood & $\bf 0.72 \pm 0.01$ & $0.66 \pm 0.01$ & $0.58 \pm 0.01$ \\
$\beta=0.1$ & $0.71 \pm 0.01$ & $\bf 0.72 \pm 0.01$ & $0.60 \pm 0.01$ \\
$\beta=0.5$ & $0.71 \pm 0.01$ & $0.69 \pm 0.01$ & $\bf 0.64 \pm 0.00$ \\
$\beta=1$ & $0.70 \pm 0.00$ & $ 0.64 \pm 0.01$ & $\bf 0.64 \pm 0.00$ \\
\hline
\end{tabular}
}
\end{table}

\textbf{Discussion:} Although Table~\ref{table:purity} demonstrates the robustness of $\beta$-GE, the difference between $\beta$-GE and existing GE is not drastic. 
This may be because the expected noise $\eta_*$ is not very different from the correct expected weight $\mu_*$.
Although $\beta$-GE may improve the likelihood-based GE drastically if some $w_{ij}$ take extremely large values as in the classical setting of robust statistics, we do not investigate such a setting in this paper.

%Thus, robustification of GE in our setting is a very challenging problem. 

\textbf{Visualization in Figure~\ref{fig:illustrative_example}:} 
Using $\xi=0.03$, feature vectors obtained by the likelihood-based GE~($\Leftrightarrow \beta=0$) and $\beta$-GE with $\beta=0.5,1.0$ are plotted in Figure~\ref{fig:illustrative_example} as an illustrative example. 
In each figure, nodes are colored by the four clusters, and links are shown as gray segments. 
Our proposed $\beta$-GE distinguishes the four clusters even when noisy links are included in the synthetic dataset, 
whereas the clusters are unclear for the likelihood-based GE.

\subsection{Experiment on the Cora Citation Network Dataset}
\label{subsec:real_experiment_1}

\textbf{Dataset:} Cora citation network~\citep{sen2008collective} consists of 2,708 nodes and 5,278 ordered edges.
Each  node $v_i$ represents  a  document,  which  has 1,433-dimensional (bag-of-words) data vector $\bs x_i \in \{0,1\}^{1,433}$ and a class label of 7 classes. Each directed edge represents citation from a document $v_i$ to another document $v_j$. 
We set the link weight as $w_{ij}=w_{ji}=1$ by ignoring the direction, and $w_{ij}=0$ otherwise. 
%There is no cross- or self-citation. この情報は不要
We divide the dataset into a training set consisting of 2,166 nodes ($80 \%$) with their edges, and a test set consisting of the remaining $542$ nodes ($20\%$) with their edges. 
Hyper-parameters are tuned by a validation set consisting of $20\%$ of the training set.

\textbf{Neural network architecture:} 
We employ a one-hidden-layer fully connected network, which consists of 3,000 tanh hidden units and 100 tanh output units.
The dimension of the feature vector is $K=100$. 
The minibatch-based stochastic algorithm shown in Section~\ref{subsec:minibatch_SGD} is used for optimization with batch normalization and dropout ($p = 0.5$).
The learning rate and momentum are tuned on the validation set.
%We monitor the score on the validation set for early stopping.

\textbf{Evaluation:} 
$k$-means clustering is applied to the obtained feature vectors of nodes representing documents.
The number of clusters is set as $7$, and the clustering result is evaluated by normalized mutual information~\cite[NMI; see][]{manning2008introduction}.
We compare the result with the
stochastic block model~\citep[SBM; see][]{holland1983stochastic}, 
ISOMAP~\citep{tenenbaum2000global}, 
locally linear embedding~\citep[LLE; see][]{roweis2000nonlinear}, 
SGE, 
multi-dimensional scaling~\citep[MDS; see][]{kruskal1964multidimensional}, 
DeepWalk~\citep{perozzi2014deepwalk}, 
and GraphSAGE~\citep{hamilton2017inductive}.

\textbf{Results:} 
The quality of clustering is evaluated by NMI. 
Sample averages and standard errors over 10 experiments are listed in Table~\ref{table:cora_clustering}. 
In experiment (A), feature vectors are computed from both the training set and test set, and they are evaluated on the test set. 
In experiment (B), feature vectors are computed from only the training set, and they are evaluated on the test set. 
SGE, MDS, and DeepWalk are not inductive, and they cannot be applied to unseen data vectors in experiment (B).

\begin{table}[htbp]
\centering
\caption{Sample average and standard error of NMIs (multiplied by 100) over 10 times experiments on Cora citation dataset. 
Higher is better. }
\label{table:cora_clustering}
\scalebox{1.0}{
\begin{tabular}{lcc}
\hline
& (A) & (B) \\
\hline
SBM$^{\dagger}$ & $4.37 \pm 0.46$ & $2.81 \pm 0.03$ \\
ISOMAP$^{\dagger}$ & $13.0 \pm 0.11$ & $14.3 \pm 0.63$ \\
LLE$^{\dagger}$ & $7.40 \pm 1.08$ & $9.47 \pm 0.95$ \\
SGE$^{\dagger}$ & $1.41 \pm 0.11$ & - \\
MDS$^{\dagger}$ & $2.81 \pm 0.03$ & - \\
DeepWalk$^{\dagger}$ & $16.7 \pm 0.33$ & - \\
GraphSAGE$^{\dagger}$ & $19.6 \pm 0.29$ & $12.4 \pm 0.95$ \\
Likelihood-based$^{\dagger}$ & $35.9 \pm 0.28$ & $30.5 \pm 1.23$ \\
\textbf{$\beta$-GE ($\beta=0.1$)} & ${\bf 36.2} \pm 0.28$ & ${\bf 30.7} \pm 1.11$ \\
\textbf{$\beta$-GE ($\beta=0.5$)} & ${\bf 36.1} \pm 0.28$ & ${\bf 31.0} \pm 0.94$ \\
\hline
%\textbf{$\beta$-GE ($\beta=1.0$)} & &$29.7 \pm 3.25$ \\
\end{tabular}
}
%\vspace{0.1em}
\begin{center}
\scalebox{0.8}{$^{\dagger}$  scores are referred to \citet{okuno2018probabilistic}.}
\end{center}
\end{table}

%
%\subsection{Experiment on Co-Authorship Network Dataset~(途中)}
%\label{subsec:real_experiment_2}
%In this section, we conduct link prediction experiment on the following dataset, to show that $\beta$-GE outperforms existing methods. 
%
%
%
%\textbf{Dynamic co-authorship network}~\citep{desmier2012cohesive} consists of 2,723 nodes and 9 layers of links. 
%Each node $v_i$ represents an author, which has $43$-dimensional data vector $\bs x_i \in \mathbb{N}_0^{43}$ representing the number of publications in each of $43$ conferences/journals. 
%Each author publishes at least 10 papers in total. 
%Link $w_{ij}$ in each layer represents the co-authorship relationship between $v_i,v_j$ at each timestamp; $w_{ij}=w_{ji}=1$ if $v_{i}$ and $v_{j}$ publishes at least one paper as co-author at the timestamp, and $w_{ij}=w_{ji}=0$ if not. 
%

\section{CONCLUSION}
\label{sec:conclusion}
We have proposed $\beta$-GE, by incorporating the newly proposed EMBS into existing likelihood-based GE. 
We have proved that $\beta$-GE is robust against noise in link weights, and is free from the distributional specification. 
We have also proposed an efficient minibatch-based stochastic algorithm that is theoretically proven to exactly locally minimize EMBS. 
Although robustification of GE is very challenging in practice, numerical experiments on synthetic and real-world datasets demonstrated the promising performance of $\beta$-GE compared with existing methods.

\section*{ACKNOWLEDGEMENT}
We would like to thank Takayuki Kawashima for helpful discussions. 
This work was partially supported by JSPS KAKENHI grant 16H02789 to HS and 17J03623 to AO.

%%% Referencesは無限ページ可能らしい(AISTATS2019)
%\bibliographystyle{apalike}
%\bibliography{robust_pmvge}

\clearpage
\onecolumn

\appendix

\begin{flushleft}
\textbf{\Large Supplementary Material:} \par
{\Large Robust Graph Embedding with Noisy Link Weights}
\end{flushleft}
\hrulefill

\section{Lemmas and Proofs}

\subsection{Law of Large Numbers for Doubly-Indexed Partially-Dependent Random Variables}
\label{appendix:mlln}
In this section, we first show and prove Theorem~\ref{theo:mlln}, that is the law of large numbers, for doubly-indexed partially-dependent random variables. Then, we apply Theorem~\ref{theo:mlln} to the empirical probability $\beta$-score and the empirical moment $\beta$-score for proving Lemma~\ref{lem:lln_beta_ce} and \ref{lem:lln_EMBS} in which we show  convergence
\begin{align*}
    (\ref{eq:empirical_beta_div}) \overset{p}{\to} E_{\mathcal{X}^2}(u_{\beta}^{(w_{12} \mid \bs x_1,\bs x_2)}(q,p_{\bs \theta})), \quad
    L_{\beta,n}(\bs \theta) \overset{p}{\to} u_{\beta}^{(\bs x_1,\bs x_2)}(g,\mu_{\bs \theta};\nu), 
\end{align*}    
as $n \to \infty$, respectively.

\begin{theo}
\label{theo:mlln}
Let $\bs Z:=(Z_{ij})$ be an array of random variables $Z_{ij} \in \mathcal{Z}$,  $(i,j) \in \mathcal{I}_n:=\{(i,j) \mid 1 \leq i < j \leq n\}$, and $h:\mathcal{Z} \to \mathbb{R}$ be a bounded and continuous function. 
We assume that $Z_{ij}$ is independent of $Z_{kl}$ if $(k,l) \in \mathcal{R}_n(i,j):=\{(k,l) \in \mathcal{I}_n \mid k,l \in \{1,\ldots, n\} \setminus \{i,j\} \}$, and 
$E_{\bs Z}(h(Z_{ij})^2)<\infty$, for all $(i,j) \in \mathcal{I}_n$. Then the average of $h(Z_{ij})$ over $\mathcal{I}_n$ converges to the expectation in probability as $n\to\infty$; that is
\begin{align*}
	\frac{1}{|\mathcal{I}_n|}\sum_{(i,j) \in \mathcal{I}_n} h(Z_{ij})
	=
	\frac{1}{|\mathcal{I}_n|}\sum_{(i,j) \in \mathcal{I}_n} E_{\bs Z}(h(Z_{ij}))
	+
	O_p(1/\sqrt{n}).	
\end{align*}
\end{theo}

\textbf{Proof of Theorem~\ref{theo:mlln}.} 
Regarding the variance of the average, we have
\begin{align*}
V_{\bs Z}\left( \frac{1}{|\mathcal{I}_n|}\sum_{(i,j) \in \mathcal{I}_n}h(Z_{ij}) \right)
&=
E_{\bs Z}\left(
	\left(\frac{1}{|\mathcal{I}_n|}\sum_{(i,j) \in \mathcal{I}_n}h(Z_{ij})\right)^2
\right)
-
E_{\bs Z}\left(
	\frac{1}{|\mathcal{I}_n|}\sum_{(i,j) \in \mathcal{I}_n}h(Z_{ij})
\right)^2 \\
&=
\frac{1}{|\mathcal{I}_n|^2}
\left(
\sum_{(i,j) \in \mathcal{I}_n}
\sum_{(k,l) \in \mathcal{I}_n}
E_{\bs Z}\left(
	h(Z_{ij})h(Z_{kl})
\right)
-
\left(
\sum_{(i,j) \in \mathcal{I}_n}
E_{\bs Z}\left(h(Z_{ij})\right)
\right)
^2
\right) \\
&=
\frac{1}{|\mathcal{I}_n|^2}
\sum_{(i,j) \in \mathcal{I}_n}
\sum_{(k,l) \in \mathcal{I}_n \setminus \mathcal{R}_n(i,j)}
\left(
	E_{\bs Z}\left(h(Z_{ij})h(Z_{kl})\right)
	-
	E_{\bs Z} \left( h(Z_{ij}) \right) E_{\bs Z} \left(h(Z_{kl})\right)
\right),
\end{align*}
where $E_{\bs Z},V_{\bs Z}$ represent expectation and variance with respect to $\bs Z$. 
By considering 
$E_{\bs Z}(|h(Z_{ij})|)\le E_{\bs Z}(h(Z_{ij})^2)^{1/2}<\infty,
E_{\bs Z}(|h(Z_{ij})h(Z_{kl})|)\le \sqrt{E_{\bs Z}(h(Z_{ij})^2)E_{\bs Z}(h(Z_{kl}))^2}<\infty$, $|\mathcal{I}_n|=O(n^2)$ and $|\mathcal{I}_n \setminus \mathcal{R}_n(i,j)|=O(n)$, 
the last formula is of order $O(n^{-4} \cdot n^{2} \cdot n)=O(n^{-1})$. Therefore, 
\begin{align}
V_{\bs Z}\left( \frac{1}{|\mathcal{I}_n|}\sum_{(i,j) \in \mathcal{I}_n}h(Z_{ij}) \right)
=
O(n^{-1}).
\label{eq:vz}
\end{align}
(\ref{eq:vz}) and Chebyshev's inequality indicate the assertion. 
\qed

The same assertion appears in Supplement~B.1 of \citet{okuno2018probabilistic}.
We note that the convergence rate is only $O_p(1/\sqrt{n})$ but not $O_p(1/\sqrt{|\mathcal{I}_n|})=O_p(1/n)$, even though we leverage $O(|\mathcal{I}_n|)=O(n^2)$ observations $\{Z_{ij}\}_{(i,j) \in \mathcal{I}_n}$. 
% This convergence rate is similar to that of U-statistic~\citep{hoeffding1948class}. 

\begin{lem}
\label{lem:lln_beta_ce}
Let $\bs \Theta$ be a parameter set. 
Assuming that $w_{ij} \mid \bs x_i,\bs x_j \overset{\text{indep.}}{\sim} q$,
$\bs x_i \overset{\text{i.i.d.}}{\sim}Q$, 
$\text{supp}Q \subset \mathcal{X}$ where $\mathcal{X}\subset \mathbb{R}^p$ is a compact set, 
$\sum_{w \in \mathbb{N}_0}q(w \mid \bs x_1,\bs x_2)p_{\bs \theta}(w \mid \bs x_1,\bs x_2)^{\delta} <\infty,
\sum_{w \in \mathbb{N}_0}p_{\bs \theta}(w \mid \bs x_1,\bs x_2)^{1+\delta} <\infty$ for all 
$\delta > 0,\bs x_1,\bs x_2 \in \mathcal{X}$. Then, it holds for all $\bs \theta \in \bs \Theta$ that
\begin{align*}
(\ref{eq:empirical_beta_div})
=
E_{\mathcal{X}^2}(d_{\beta}^{(w_{12} \mid \bs x_1,\bs x_2)}(q,p_{\bs \theta})) 
+
O_p(1/\sqrt{n}),
\end{align*}
indicating 
$(\ref{eq:empirical_beta_div})
\overset{p}{\to} 
E_{\mathcal{X}^2}(d_{\beta}^{(w_{12} \mid \bs x_1,\bs x_2)}(q,p_{\bs \theta})) \: (n \to \infty)$.
\end{lem}

\textbf{Proof of Lemma~\ref{lem:lln_beta_ce}.} 
Applying Theorem~\ref{theo:mlln} to 
\begin{align*}
Z_{ij}:=(w_{ij},\bs x_i,\bs x_j), \:
h(Z_{ij}):=-\frac{p_{\bs \theta}(w_{ij} \mid \bs x_i,\bs x_j)^{\beta}-1}{\beta}
+
\sum_{w \in \mathbb{N}_0}
\frac{p_{\bs \theta}(w \mid \bs x_i,\bs x_j)^{1+\beta}}{1+\beta},
\end{align*}
% $Z_{ij}$ is obviously independent of $Z_{kl}$ if $(k,l) \in \mathcal{R}_n(i,j)$ for all $(i,j) \in \mathcal{I}_n$. 
immediately proves the assertion, as $E_{\bs Z}(h(Z_{ij})^2)<\infty$ follows from the assumptions; the convergence limit is, 
\begin{align*}
% (\ref{eq:empirical_beta_div})
% &=
% \frac{1}{|\mathcal{I}_n|}
% \sum_{(i,j) \in \mathcal{I}_n}
% h(Z_{ij}) \\
% &\overset{\text{Theorem~\ref{theo:mlln}}}{=}
\frac{1}{|\mathcal{I}_n|}&
\sum_{(i,j) \in \mathcal{I}_n}
E_{\bs Z}(h(Z_{ij})) \\
&=
\frac{1}{|\mathcal{I}_n|}
\sum_{(i,j) \in \mathcal{I}_n}
E_{\mathcal{X}^2}\left(
E\left(
-\frac{p_{\bs \theta}(w_{ij} \mid \bs x_i,\bs x_j)^{\beta}-1}{\beta}
+
\sum_{w \in \mathbb{N}_0}
\frac{p_{\bs \theta}(w \mid \bs x_i,\bs x_j)^{1+\beta}}{1+\beta}
\bigg| \bs x_i,\bs x_j\right)
\right) \\
&=
\frac{1}{|\mathcal{I}_n|}
\sum_{(i,j) \in \mathcal{I}_n}
E_{\mathcal{X}^2}\left(
\sum_{w' \in \mathbb{N}_0} 
q(w' \mid \bs x_i,\bs x_j) \left\{
-\frac{p_{\bs \theta}(w' \mid \bs x_i,\bs x_j)^{\beta}-1}{\beta}
+
\sum_{w \in \mathbb{N}_0}
\frac{p_{\bs \theta}(w \mid \bs x_i,\bs x_j)^{1+\beta}}{1+\beta}
\right\} 
\right) \\
&=
\frac{1}{|\mathcal{I}_n|}
\sum_{(i,j) \in \mathcal{I}_n}
E_{\mathcal{X}^2}\left(
-\sum_{w \in \mathbb{N}_0} 
q(w \mid \bs x_i,\bs x_j) \frac{p_{\bs \theta}(w \mid \bs x_i,\bs x_j)^{\beta}-1}{\beta}
+
\sum_{w \in \mathbb{N}_0} 
\frac{p_{\bs \theta}(w \mid \bs x_i,\bs x_j)^{1+\beta}}{1+\beta}
\right) \\
&=
\frac{1}{|\mathcal{I}_n|}
\sum_{(i,j) \in \mathcal{I}_n}
E_{\mathcal{X}^2}(d_{\beta}^{(w_{ij} \mid \bs x_i,\bs x_j)}(q,p_{\bs \theta})) \\
&=
\frac{1}{|\mathcal{I}_n|}
\sum_{(i,j) \in \mathcal{I}_n}
E_{\mathcal{X}^2}(d_{\beta}^{(w_{12} \mid \bs x_1,\bs x_2)}(q,p_{\bs \theta})) \\
&=
E_{\mathcal{X}^2}(d_{\beta}^{(w_{12} \mid \bs x_1,\bs x_2)}(q,p_{\bs \theta})).
\end{align*}
% \end{enumerate}
Thus proving the assertion. 
\qed

\begin{lem}
\label{lem:lln_EMBS}
Let $\bs \Theta$ be a parameter set. 
Assuming (\ref{eq:w_cont_model})--(\ref{eq:x_model}), it holds for all $\bs \theta \in \bs \Theta$ that
\begin{align*}
L_{\beta,n}(\bs \theta)
=
u_{\beta}^{(\bs x_1,\bs x_2)}
(g,\mu_{\bs \theta};\nu)
+
O_p(1/\sqrt{n}),
\end{align*}
indicating 
$L_{\beta,n}(\bs \theta)
\overset{p}{\to} 
u_{\beta}^{(\bs x_1,\bs x_2)}
(g,\mu_{\bs \theta};\nu) \: (n \to \infty)$.
\end{lem}
% We note that $\mu_{\bs \theta}(\bs x_1,\bs x_2)$ is defined to be continuous with respect to $(\bs x_1,\bs x_2) \in \mathcal{X}$.

\textbf{Proof of Lemma~\ref{lem:lln_EMBS}.} 
Applying Theorem~\ref{theo:mlln} to
\begin{align*}
Z_{ij}:=(w_{ij},\bs x_i,\bs x_j), \:
h(Z_{ij}):=-w_{ij} \frac{\mu_{\bs \theta}(\bs x_i,\bs x_j)^{\beta}-1}{\beta}
+
\frac{\mu_{\bs \theta}(\bs x_i,\bs x_j)^{1+\beta}}{1+\beta},
\end{align*}
% $Z_{ij}$ is obviously independent of $Z_{kl}$ if $(k,l) \in \mathcal{R}_n(i,j)$ for all $(i,j) \in \mathcal{I}_n$. 
immediately proves the assertion, as $E_{\bs Z}(h(Z_{ij})^2)<\infty$ follows from the assumptions; the convergence limit is,

\begin{align*}
% L_{\beta,n}(\bs \theta)
% &=
% \frac{1}{|\mathcal{I}_n|}
% \sum_{(i,j) \in \mathcal{I}_n}
% h(Z_{ij}) \\
% &\overset{\text{Theorem~\ref{theo:mlln}}}{=}
\frac{1}{|\mathcal{I}_n|}
\sum_{(i,j) \in \mathcal{I}_n}
E_{\bs Z}(h(Z_{ij})) 
&=
\frac{1}{|\mathcal{I}_n|}
\sum_{(i,j) \in \mathcal{I}_n}
E_{\mathcal{X}^2}\left(
	E\left(
		-w_{ij} \frac{\mu_{\bs \theta}(\bs x_i,\bs x_j)^{\beta}-1}{\beta}
+
\frac{\mu_{\bs \theta}(\bs x_i,\bs x_j)^{1+\beta}}{1+\beta}
\bigg| \bs x_i,\bs x_j
	\right)
\right) \\
&=
\frac{1}{|\mathcal{I}_n|}
\sum_{(i,j) \in \mathcal{I}_n}
E_{\mathcal{X}^2}\left(
		-g(\bs x_i,\bs x_j) \frac{\mu_{\bs \theta}(\bs x_i,\bs x_j)^{\beta}-1}{\beta}
+
\frac{\mu_{\bs \theta}(\bs x_i,\bs x_j)^{1+\beta}}{1+\beta}
\right) \\
&=
\frac{1}{|\mathcal{I}_n|}
\sum_{(i,j) \in \mathcal{I}_n}
u_{\beta}^{(\bs x_i,\bs x_j)}(g,\mu_{\bs \theta};\nu) \\
&=
\frac{1}{|\mathcal{I}_n|}
\sum_{(i,j) \in \mathcal{I}_n}
u_{\beta}^{(\bs x_1,\bs x_2)}(g,\mu_{\bs \theta};\nu) \\
&=
u_{\beta}^{(\bs x_1,\bs x_2)}(g,\mu_{\bs \theta};\nu).
\end{align*}
% \end{enumerate}
% As $n \to \infty$, the term $O_p(1/\sqrt{n})$ goes to $0$. 
Thus proving the assertion. \qed

\subsection[Evaluation of a Term in Main Theorem]{Evaluation of $M(\bs \theta)$ in Theorem~\ref{theo:div_perturbation}}

\begin{lem}
\label{lem:order_epsilon}
Suppose that $\varepsilon \geq \varepsilon_*$, $\bs \theta \in \bs \Theta_{\varepsilon}:=\{
    \bs \theta \in \bs \Theta \mid 
    E_{\mathcal{X}^2}(\eta_*(\bs x_1,\bs x_2)\mu_{\bs \theta}(\bs x_1,\bs x_2)^{\beta_0})<\varepsilon
\}$, and $\beta \in (0,\beta_0]$, it holds for
\[
M (\bs \theta) := \beta^{-1}
    E_{\mathcal{X}^2}
\left(
    \eta_*(\bs x_1,\bs x_2)
    \mu_{\bs \theta}(\bs x_1,\bs x_2)^{\beta}
\right)\varepsilon^{-\beta/\beta_0},
\quad
\alpha:=E_{\mathcal{X}^2}(\eta_*(\bs x_1,\bs x_2)),
\]
that 
\begin{align*}
    M(\bs \theta)
    &\leq
    \alpha^{1-\beta/\beta_0}\beta^{-1}
    \quad (\forall \bs \theta \in \bs \Theta_{\varepsilon}).
\end{align*}
\end{lem}

\textbf{Proof of Lemma~\ref{lem:order_epsilon}.} 
Proof is based on Lyapunov's inequality, that is, $E(Z^{\beta}) \leq E(Z^{\beta_0})^{\beta/\beta_0}$ for any non-negative real-valued random variable $Z$ and $0<\beta \leq \beta_0<\infty$. 
For applying this inequality, we first fix $\bs \theta \in \bs \Theta_{\varepsilon}$, and expand $M(\bs \theta)$ with the probability density function~(pdf) $\nu$ of the random variable $(\bs x_1,\bs x_2)$ as
\begin{align}
M(\bs \theta)
&=
\beta^{-1}
E_{\mathcal{X}^2}\left(
    \eta_*(\bs x_1,\bs x_2)
    \mu_{\bs \theta}(\bs x_1,\bs x_2)^{\beta}
\right)
\varepsilon^{-\beta/\beta_0}
\nonumber \\
&=
\beta^{-1}
\varepsilon^{-\beta/\beta_0}
\iint_{\mathcal{X}^2} 
    \nu(\bs x_1,\bs x_2)
    \eta_*(\bs x_1,\bs x_2)
    \mu_{\bs \theta}(\bs x_1,\bs x_2)^{\beta}
    \mathrm{d} \bs x_1
    \mathrm{d} \bs x_2 \nonumber \\ 
&=
\alpha
\beta^{-1}
\varepsilon^{-\beta/\beta_0}
\left(
\iint_{\mathcal{X}^2} 
\underbrace{
    \frac{
    \nu(\bs x_1,\bs x_2)
    \eta_*(\bs x_1,\bs x_2)
    }{\alpha}
}_{=:\tilde{\nu}(\bs x_1,\bs x_2)}
    \mu_{\bs \theta}(\bs x_1,\bs x_2)^{\beta}
    \mathrm{d} \bs x_1
    \mathrm{d} \bs x_2 
\right). \label{eq:epsilon_beta_rhs}
\end{align}
In eq.~(\ref{eq:epsilon_beta_rhs}), $\tilde{\nu}(\bs x_1,\bs x_2):= \nu(\bs x_1,\bs x_2)
    \eta_*(\bs x_1,\bs x_2)
    /\alpha$ can be regarded as a pdf, 
    since $\tilde{\nu}(\bs x_1,\bs x_2)\geq 0$ for all $(\bs x_1,\bs x_2)$ and 
\begin{align*}
\iint_{\mathcal{X}^2}\tilde{\nu}_*(\bs x_1,\bs x_2) \mathrm{d} \bs x_1 \mathrm{d}\bs x_2
&=
\iint_{\mathcal{X}^2}\frac{\nu(\bs x_1,\bs x_2) \eta_*(\bs x_1,\bs x_2)}{\alpha} \mathrm{d} \bs x_1 \mathrm{d}\bs x_2 \\
&=
\alpha^{-1}
\iint_{\mathcal{X}^2} \nu(\bs x_1,\bs x_2) \eta_*(\bs x_1,\bs x_2) \mathrm{d} \bs x_1 \mathrm{d}\bs x_2 \\
&=
\alpha^{-1}
E_{\mathcal{X}^2}(\eta_*(\bs x_1,\bs x_2))
=
\alpha^{-1} \alpha
=
1.
\end{align*}
As $\tilde{\nu}$ can be regarded as a pdf and $\mu_{\bs \theta}$ is non-negative, Lyapunov's inequality indicates that
\begin{align}
M(\bs \theta)
=
(\ref{eq:epsilon_beta_rhs})
&\overset{\text{(Lyapunov)}}{\leq}
\alpha
\beta^{-1}
\varepsilon^{-\beta/\beta_0}
\left(
\iint_{\mathcal{X}^2} 
    \tilde{\nu}(\bs x_1,\bs x_2)
    \mu_{\bs \theta}(\bs x_1,\bs x_2)^{\beta_0}
    \mathrm{d} \bs x_1
    \mathrm{d} \bs x_2
\right)^{\beta/\beta_0} \nonumber \\
&=
\alpha
\beta^{-1}
\varepsilon^{-\beta/\beta_0}
\left(
\iint_{\mathcal{X}^2} 
\frac{
    \nu(\bs x_1,\bs x_2)
    \eta_*(\bs x_1,\bs x_2)
}{\alpha}
    \mu_{\bs \theta}(\bs x_1,\bs x_2)^{\beta_0}
    \mathrm{d} \bs x_1
    \mathrm{d} \bs x_2
\right)^{\beta/\beta_0} \nonumber \\
&=
\alpha^{1-\beta/\beta_0}
\beta^{-1}
\varepsilon^{-\beta/\beta_0}
\left(
\iint_{\mathcal{X}^2} 
    \nu(\bs x_1,\bs x_2)
    \eta_*(\bs x_1,\bs x_2)
    \mu_{\bs \theta}(\bs x_1,\bs x_2)^{\beta_0}
    \mathrm{d} \bs x_1
    \mathrm{d} \bs x_2
\right)^{\beta/\beta_0}
\nonumber \\
&=
\alpha^{1-\beta/\beta_0}
\beta^{-1}
\varepsilon^{-\beta/\beta_0}
E_{\mathcal{X}^2} 
\left(
    \eta_*(\bs x_1,\bs x_2)
    \mu_{\bs \theta}(\bs x_1,\bs x_2)^{\beta_0}
\right)^{\beta/\beta_0} \nonumber \\
&\leq
\alpha^{1-\beta/\beta_0}
\beta^{-1}
\varepsilon^{-\beta/\beta_0}
\varepsilon^{\beta/\beta_0} 
\qquad (\because \bs \theta \in \bs \Theta_{\varepsilon})
\nonumber \\
&=
\alpha^{1-\beta/\beta_0}
\beta^{-1}. \nonumber
%\label{eq:M_beta_upper_bound}
\end{align}
The assertion is proved. 
\qed

\subsection{Proof of Theorem~\ref{theo:minibatchSGD_limit}}
\label{proof_minibatchSGD}

We first verify that (\ref{eq:minibatch_estimating}) is equivalent to $\partial h(\bs\theta)/\partial \bs \theta = \bs 0$.
From the definition of $h^{(t)}(\bs \theta)$ and the assumption (i) $\mu_{\bs \theta}(\bs x_1,\bs x_2) \in C^1(\bs \Theta)$ for all $(\bs x_1,\bs x_2) \in \mathcal{X}^2$, we have
\begin{align*}
\frac{\partial h(\bs \theta)}{\partial \bs \theta}
&=
\frac{\partial E^{(1)}(h^{(1)}(\bs \theta))}{\partial \bs \theta}
=
E^{(1)}\left( \frac{\partial h^{(1)}(\bs \theta)}{\partial \bs \theta} \right) \\
&=
E^{(1)}
\bigg(
\frac{\partial}{\partial \bs \theta}
\bigg\{
    -\sum_{(i,j) \in \mathcal{W}_n^{(1)}}w_{ij} \frac{\mu_{\bs \theta}(\bs x_i,\bs x_j)^{\beta}-1}{\beta}
    +\lambda \sum_{(i,j) \in \mathcal{I}_n^{(1)}} \frac{\mu_{\bs \theta}(\bs x_i,\bs x_j)^{1+\beta}}{1+\beta}
\bigg\}
\bigg) \\
%%============================================
&=
E^{(1)}
\bigg(
\bigg\{
    -
    \sum_{(i,j) \in \mathcal{W}_n^{(1)}}w_{ij} 
    \mu_{\bs \theta}(\bs x_i,\bs x_j)^{\beta}
    \frac{\partial \log \mu_{\bs \theta}(\bs x_i,\bs x_j)}{\partial \bs \theta}
    +
    \lambda \sum_{(i,j) \in \mathcal{I}_n^{(1)}} 
    \mu_{\bs \theta}(\bs x_i,\bs x_j)^{1+\beta}
    \frac{\partial \log \mu_{\bs \theta}(\bs x_i,\bs x_j)}{\partial \bs \theta}
\bigg\}
\bigg) \\
%%============================================
&=
-
E^{(1)}
\bigg(
    \sum_{(i,j) \in \mathcal{W}_n^{(1)}}w_{ij} 
    \mu_{\bs \theta}(\bs x_i,\bs x_j)^{\beta}
    \frac{\partial \log \mu_{\bs \theta}(\bs x_i,\bs x_j)}{\partial \bs \theta}
\bigg)
    +
    \lambda 
E^{(1)}
\bigg(
    \sum_{(i,j) \in \mathcal{I}_n^{(1)}} 
    \mu_{\bs \theta}(\bs x_i,\bs x_j)^{1+\beta}
    \frac{\partial \log \mu_{\bs \theta}(\bs x_i,\bs x_j)}{\partial \bs \theta}
\bigg) \\
&=
-
\frac{m_1}{|\mathcal{W}_n|}  
    \sum_{(i,j) \in \mathcal{W}_n} w_{ij} 
    \mu_{\bs \theta}(\bs x_i,\bs x_j)^{\beta}
    \frac{\partial \log \mu_{\bs \theta}(\bs x_i,\bs x_j)}{\partial \bs \theta}
+
\lambda \frac{m_2}{|\mathcal{I}_n|} 
    \sum_{(i,j) \in \mathcal{I}_n} \mu_{\bs \theta}(\bs x_i,\bs x_j)^{1+\beta}
\frac{\partial \log \mu_{\bs \theta}(\bs x_i,\bs x_j)}{\partial \bs \theta} \\
&=
\frac{1}{|\mathcal{I}_n|}
\sum_{(i,j) \in \mathcal{I}_n}
\bigg\{
\bigg(
    - v m_1 w_{ij} 
    + \lambda m_2 \mu_{\bs \theta}(\bs x_i,\bs x_j)
\bigg) \mu_{\bs \theta}(\bs x_i,\bs x_j)^{\beta}
\frac{\partial \log \mu_{\bs \theta}(\bs x_i,\bs x_j)}{\partial \bs \theta}
\bigg\}.
\end{align*}

We next verify the convergence $E^*(\|\bs \theta^{(t)}-\bs \theta_*\|_2^2) \to 0$.
From the assumption (ii), $\bs\theta_*$ is the unique minimizer of $h(\bs\theta)$ over $\bs \Theta$.
Regarding the estimator $\bs \theta^{(t)}$ defined as (\ref{eq:minibatch_sgd_formulation}) with the assumption (iii),
\citet{eric2011non} Theorem~2 asserts that $E^*(\|\bs \theta^{(t)}-\bs \theta_*\|_2^2) \to 0$ if
the following conditions (C-1)--(C-3) hold:  
(C-1) $E^{(t)}\left( \frac{\partial h^{(t)}(\bs \theta)}{\partial \bs \theta} \right)=\frac{\partial h(\bs \theta)}{\partial \bs \theta}$ for all $\bs \theta \in \bs \Theta$, 
(C-2) $h(\bs \theta)$ is strongly convex on $\bs \Theta$, i.e., $\exists \lambda>0$ such that $ h(\bs \theta_1) -  h(\bs \theta_2) \ge \langle \frac{\partial h(\bs \theta_2)}{\partial \bs \theta},\bs \theta_1-\bs \theta_2 \rangle + \lambda \|\bs \theta_1-\bs \theta_2\|_2^2$ for all $\bs \theta_1,\bs \theta_2 \in \bs \Theta$, and
(C-3) $\|\frac{\partial h^{(t)}(\bs \theta)}{\partial \bs \theta}\|_2$ is bounded on $\bs \Theta$ for any $(\mathcal{W}_n^{(t)},\mathcal{I}_n^{(t)})$. 
These conditions (C-1)--(C-3) correspond to the conditions (H1), (H3), and (H5), that are required in \citet{eric2011non} Theorem~2, respectively.

In case of Theorem~\ref{theo:minibatchSGD_limit}, 
(C-1) holds as we have already seen for showing  (\ref{eq:minibatch_estimating}); note that $h^{(t)}(\bs \theta) \in C^1(\bs \Theta)$ from the assumption (i).
(C-2) is assumed as (ii), 
and 
(C-3) holds because $h^{(t)}(\bs \theta)$ is $C^1$ on the compact set $\bs \Theta$ and $(\mathcal{W}^{(t)}_n,\mathcal{I}^{(t)}_n)$ is a random variable taking value in a finite set. Thus we have proved the convergence.

 \qed

\end{document}